\documentclass[manuscript,screen]{acmart}

\usepackage{flushend}
\usepackage{caption}
\usepackage{tabularx}
\usepackage{graphicx}
\usepackage{subfigure}
\usepackage{amsfonts}
\usepackage{amsmath}
\usepackage{amsthm}
\usepackage{algorithm}
\usepackage{algorithmic}
\usepackage{multirow}
\usepackage{booktabs}
\usepackage{float}
\usepackage{bm}
\usepackage{xspace}
\usepackage{threeparttable}
\usepackage{hyperref}
\usepackage{url}
\usepackage{mathtools}
\usepackage{makecell}
\usepackage{subcaption}
\usepackage{color}
\usepackage{xcolor}
\usepackage{balance}
\usepackage{enumitem}

\usepackage{amssymb}

\newcommand{\ie}{\textit{i}.\textit{e}.}
\newcommand{\eg}{\textit{e}.\textit{g}.} 

\newcommand{\wrt}{\textit{w}.\textit{r}.\textit{t}}

\newcommand{\model}{MHCL\xspace}

\AtBeginDocument{%
  }

\setcopyright{acmlicensed}
\copyrightyear{2025}
\acmYear{2025}
\acmDOI{XXXXXXX.XXXXXXX}
\acmConference[Conference acronym 'XX]{Make sure to enter the correct
  conference title from your rights confirmation email}{June 03--05,
  2018}{Woodstock, NY}
\acmISBN{978-1-4503-XXXX-X/2018/06}

\begin{document}

\title{Multi-Channel Hypergraph Contrastive Learning for Matrix Completion}

\author{Xiang Li}
\authornote{Xiang Li and Changsheng Shui contribute equally to this work.}
\affiliation{%
  \institution{Ocean University of China}
  \city{Qingdao}
  \country{China}}
\email{lixiang1202@stu.ouc.edu.cn}

\author{Changsheng Shui}
\authornotemark[1]
\affiliation{
  \institution{Ocean University of China}
  \city{Qingdao}
  \country{China}
}
\email{scs@stu.ouc.edu.cn}


\author{Zhongying Zhao}
\affiliation{
  \institution{Shandong University of Science and Technology}
  \city{Qingdao}
  \country{China}
}
\email{zyzhao@sdust.edu.cn}

\author{Junyu Dong}
\affiliation{%
  \institution{Ocean University of China}
  \city{Qingdao}
  \country{China}
}
\email{dongjunyu@ouc.edu.cn}

\author{Yanwei Yu}
\authornote{Yanwei Yu is the corresponding author.}
\affiliation{%
  \institution{Ocean University of China}
  \city{Qingdao}
  \country{China}}
\email{yuyanwei@ouc.edu.cn}

\renewcommand{\shortauthors}{Li and Shui et al.}


\begin{abstract}
Rating is a typical user's explicit feedback that visually reflects how much a user likes a related item. The (rating) matrix completion is essentially a rating prediction process, which is also a significant problem in recommender systems. Recently, graph neural networks (GNNs) have been widely used in matrix completion, which captures users’ preferences over items by formulating a rating matrix as a bipartite graph. However, existing methods are susceptible due to data sparsity and long-tail distribution in real-world scenarios. Moreover, the messaging mechanism of GNNs makes it difficult to capture high-order correlations and constraints between nodes, which are essentially useful in recommendation tasks. To tackle these challenges, we propose a \textbf{\underline{M}}ulti-Channel \textbf{\underline{H}}ypergraph \textbf{\underline{C}}ontrastive \textbf{\underline{L}}earning framework for matrix completion, named \textbf{MHCL}. Specifically, MHCL adaptively learns hypergraph structures to capture high-order correlations between nodes and jointly captures local and global collaborative relationships through attention-based cross-view aggregation. Additionally, to consider the magnitude and order information of ratings, we treat different rating subgraphs as different channels, encourage alignment between adjacent ratings, and further achieve the mutual enhancement between different ratings through multi-channel cross-rating contrastive learning. Extensive experiments on eight publicly available real-world datasets demonstrate that our proposed method significantly outperforms the current state-of-the-art approaches. The source code of our model is available at \url{https://anonymous.4open.science/r/MHCL-25DC}.
\end{abstract}



\received{20 March 2025}
\received[revised]{20 August 20025}
\received[accepted]{20 December 2025}

\maketitle

\section{Introduction}  
In recent years, with the explosion of information, recommender systems have become an indispensable tool for e-commerce and social media platforms~\cite{zhang2019star, ricci2010introduction, xia2020multiplex}. In recommender systems, users' feedback on items is divided into explicit and implicit feedback~\cite{chen2022review,li2025dual}. Implicit feedback means that the user does not explicitly like or dislike the item, which reveals that this type of data is generally considered positive feedback for all interactions~\cite{palomares2018multi, raza2019progress}. For example, click records, shopping carts of shopping sites, and favorites or forwarding of items are part of the implicit feedback. Explicit feedback reflects how much a user likes a related item. In particular, given a partially observed matrix of users to items, the values in the matrix represent the users' ratings of items. Then the process of completing the matrix can be regarded as the prediction process of ratings.

As a kind of traditional matrix completion method, matrix factorization~\cite{candes2012exact, kalofolias2014matrix, xu2013speedup} aims to decompose the user-item rating matrix into the user latent factors and item latent factors that profile users and items accurately. Recently, Graph Neural Networks (GNNs) have been widely used in recommender systems~\cite{monti2017geometric, shen2021inductive, zhang2019inductive, zhang2022geometric}, these GNN-based methods formulate the rating matrix as a bipartite graph, and ratings in the matrix are treated as different types of edges, thus transforming the matrix completion problem into a link prediction~\cite{martinez2016survey,kumar2020link,arrar2024comprehensive} problem on the bipartite graph. Researchers introduce inductive matrix completion methods, which can easily generalize to the nodes unseen during training and enhance the generalization of the model. IGMC~\cite{zhang2019inductive} extracts a 1-hop subgraph around all target edges, relabels nodes as the initial features in each subgraph, and applies GNN to each subgraph to learn the local graph pattern that can be generalized to the invisible graphs. GIMC~\cite{zhang2022geometric} introduces edge embeddings to model the semantic properties of different types of explicit feedback based on extracting subgraphs, and further improves the modeling ability of bipartite graphs by introducing hyperbolic geometry as the vertex-level embedding space. In addition, there are some review-based models~\cite{wu2019reviews, gao2020set, shuai2022review} that utilize the auxiliary review information accompanied with user ratings to enrich the embedding learning.

\begin{figure*}[htbp]
\begin{center}
\includegraphics[scale=0.33]{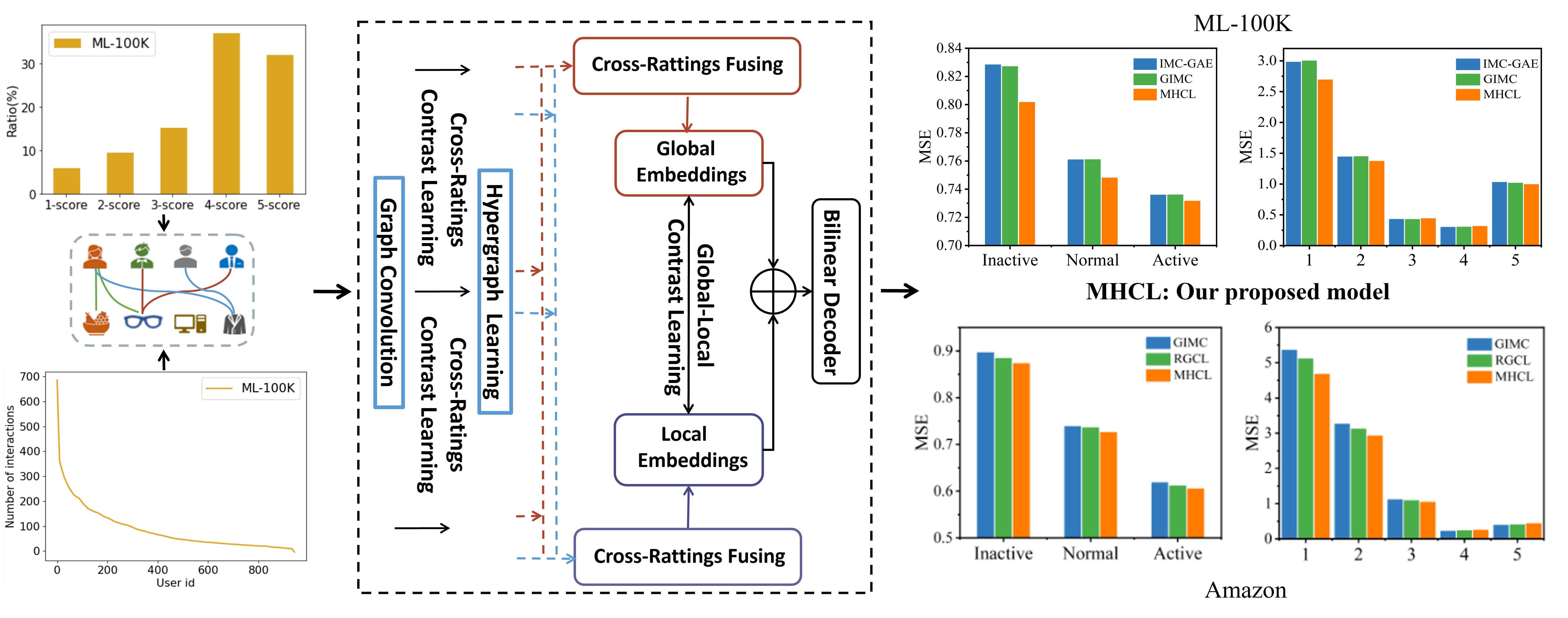}
\caption{The imbalance in the number of interactions between different nodes and rating imbalance caused by the long-tail distribution problem. Specifically, we designate 80\% of users with the fewest interactions as `Inactive', classify the top 5\% of users with the most interactions as `Active', and assign the remaining users as `Normal'. Then we calculate the mean square error for each interaction group and rating group.} 
\label{fig:intro}
\end{center}
\end{figure*}

Despite the remarkable success, existing GNN-based matrix completion methods still suffer from two significant challenges: \textit{First, existing GNN-based methods struggle to capture high-order correlations and complex constraints between nodes}. GNN-based methods rely on existing edges in the bipartite graph to learn the node representations. Still, for high-order correlations and constraints, traditional GNNs need to stack multiple layers to capture, which not only brings much noise but makes it difficult to mitigate over-smoothing issues~\cite{min2020scattering}. \textit{Second, user-item interaction data is usually sparse and follows long-tail and power-law distributions, while GNN-based methods are inherently vulnerable to data sparsity and long-tail distributions}. The long-tail distribution problem in recommender systems~\cite{mcnee2006being} refers to the imbalance in the number of interactions between different nodes; that is, a few highly active nodes dominate interactions, resulting in popularity bias and diminished effectiveness for less active nodes. For rating recommendations, the long-tail distribution will also lead to a rating imbalance. Experimental results in~\cite{mahadevan2021class, wei2021towards, zhao2020improving} show that the imbalance distribution will cause popularity bias~\cite{wu2022gumbel}, which will lead to the model being more biased towards active nodes and ratings, and reduce the recommendation effect of inactive nodes and ratings. In Figure~\ref{fig:intro}, we illustrate the long-tail distribution problem through Amazon dataset, which has the problem of the imbalance in the number of interactions and rating imbalance. We can see that the existing models (IMC-GAE~\cite{shen2021inductive} and GIMC~\cite{zhang2022geometric} as examples) have significantly higher prediction errors for tail classes (\ie, 1-score, 2-score, `Inactive' group) than for head classes (\ie, 3-score, 4-score, 5-score, `Normal' and `Active' group), while our model can reduce the prediction error for tail classification.
Unlike implicit feedback, rating information is based on the magnitude and order (rating `4' is closer to `5' rather than `1'), which may be ignored by many matrix completion methods. Although there are some recent methods~\cite{zhang2019inductive, shen2021inductive} that use regularization to mitigate these problems, they still do not consider these from a model perspective.

Desired by addressing the challenges mentioned above, we propose a novel \textbf{\underline{M}}ulti-Channel \textbf{\underline{H}}ypergraph \textbf{\underline{C}}ontrastive \textbf{\underline{L}}earning framework for the matrix completion problem, named \textbf{MHCL}. First, we divide the bipartite graph into different subgraphs according to different types of rating edges through subgraph partition and graph convolution module, and then dynamically learn hypergraph structures for different subgraphs at different layers during the training process through an adaptive hypergraph structure learning module, to solve the first challenge, to capture the high-order correlations between nodes. Second, we propose a multi-channel cross-rating contrastive learning module, which treats different rating subgraphs as different (ratings) channels and achieves mutual enhancement by encouraging the alignment between adjacent views. Third, we also design an attention-based cross-view fusing module to further enhance the interaction between different ratings to mitigate the data sparsity and long-tail distribution challenge. Finally, we use an attention-based cross-view aggregation module to collaboratively learn node representations from local and global perspectives and fuse them to obtain the final representation. Experimental results demonstrate that MHCL achieves a +9.04\% performance lift in terms of MSE compared to state-of-the-art baselines for matrix completion on ML-100K dataset. Besides, MHCL consistently outperforms advanced rating and review-based recommendation baselines on recommendation tasks, with the most notable +10.31\% N@10 improvement on ML-100K and average +6.65\% improvement on M@10 and N@10 across all datasets.

To sum up, this paper makes the contributions as follows:
\begin{itemize}[leftmargin=*]
    \item MHCL introduces a multi-channel cross-rating contrastive learning approach, treating each rating subgraph as a rating channel, which strengthens the alignment between adjacent views and improves the modeling of inter-rating correlations. This approach extracts richer learning signals, improving the representations of less frequent ratings.
    \item MHCL captures high-order correlations by dynamically learning hypergraph structures during training, effectively mitigating over-smoothing issues and enhancing the model’s ability to model complex correlations.
    \item To mitigate data sparsity and long-tail distributions, MHCL employs an attention-based cross-view fusion module, which enhances interactions between rating subgraphs and ensures better representation of both active and inactive nodes. Moreover, we integrate contrastive learning modules to holistically capture cross-rating intrinsic correlations.
    \item Extensive experiments demonstrate that MHCL significantly outperforms state-of-the-art baselines in both matrix completion and recommendation tasks, proving its effectiveness in scenarios with sparse and long-tail data distributions.
\end{itemize}

\section{Related Work}
\subsection{GNNs for Matrix Completion}
Recently, the field of graph neural networks (GNNs) has demonstrated remarkable efficacy in the domain of matrix completion. The fundamental concept underlying GNN-based matrix completion methods is the interpretation of a matrix as a bipartite graph, where observed ratings or purchases are denoted by connecting edges, which ingeniously and effectively transforms the matrix completion problem into a predictive task centered around graph edges. Traditional GNN-based methods~\cite{kipf2016semi,hamilton2017inductive,velivckovic2017graph} acquire
the information from the neighbors in the user-item graph through multi-hop convolutions to exploit the high-order relations defined by the initial pairwise links. NGCF~\cite{wang2019neural} and PinSage~\cite{ying2018graph} adopt multiple graph convolutions to explore high-order connectivity. 
LightGCN~\cite{he2020lightgcn} demonstrates that neighborhood aggregation is the most essential component in GCN and proposes to omit the non-linear transformation during propagation. IDCF-NN~\cite{wu2021towards} and IDCF-GC~\cite{wu2021towards} are both inductive collaborative filtering methods; the former adopts the traditional matrix factorization method to factorize the rating matrix of a set of key users to obtain meta-latent factors, and the latter estimates the hidden relationship from query to key users and learns to use meta-latents to inductively compute the embedding of query users through neural information transfer. GC-MC~\cite{berg2017graph} adopts a one-hot encoding scheme for node IDs as initial node features. It refines this representation through a GNN encoder applied to the bipartite graph and subsequently reconstructs the rating edges by employing a GNN decoder. Additionally, IMC-GAE~\cite{shen2021inductive} leverages identical features, role-aware attributes, and one-hot indices as initial features, enhancing the model's generalization capabilities. In a similar vein, IGMC~\cite{zhang2019inductive} introduces a novel strategy by extracting enclosing subgraphs surrounding target edges. It employs GNNs to learn local graph patterns, enabling inductive matrix completion that generalizes to users or items not encountered during training. Taking a step further, GIMC~\cite{zhang2022geometric} extends inductive matrix completion into non-Euclidean spaces by incorporating hyperbolic regression. This innovation effectively mitigates the challenges posed by data sparsity. RMG~\cite{wu2019reviews} represents a pioneering effort, amalgamating graph signals and review information for recommender systems. Subsequently, RGCL~\cite{shuai2022review} builds upon this foundation by harnessing the semantic information embedded within review features. It fine-tunes the influence of neighboring nodes and employs contrastive learning techniques to achieve superior interaction modeling. Shuai et al.~\cite{shuai2023topic} propose a novel framework TGNN, which enhances explainable recommendation by using topic modeling and GNNs to explicitly and implicitly model user preferences from reviews. This overcomes the limitations of sparsity and attribute dependency, improving both accuracy and explanation quality. MAGCL~\cite{wang2023multi} uses multi-view representation learning, graph contrastive learning and multi-task learning to capture fine-grained semantic information in user reviews, and learns higher-order structure-aware and self-discriminative node embeddings.

In summary, these advancements within the realm of GNNs have ushered in transformative approaches to matrix completion, showcasing their potential to revolutionize recommender systems.

\subsection{Contrastive Learning} 
Contrastive learning has emerged as a potent self-supervised framework, and the main idea of self-supervised contrastive learning is to encourage the representation affinity among various views of the same object, while at the same time enlarging the representation dispersion of different objects. The fundamental premise of contrastive learning revolves around bringing an ``anchor'' and a ``positive'' sample closer in the embedding space while simultaneously pushing the anchor away from multiple ``negative'' samples~\cite{ma2022crosscbr}. Drawing inspiration from the successes of contrastive learning across various domains~\cite{chen2020simple, giorgi2020declutr, oord2018representation, cao2021grammatical,yuan2021multimodal,lan2019albert}, numerous researchers have incorporated contrastive learning techniques into graph data to offer a potential solution to the challenges of insufficient supervision signals in recommender systems~\cite{velivckovic2018deep,zhu2020deep,you2020graph,thakoor2021bootstrapped,ren2019heterogeneous}.
SGL~\cite{wu2021self}, for instance, enriches the user-item bipartite graph with structural perturbations and subsequently focuses on maximizing the consistency of representations derived from different perspectives, as learned through a graph encoder. SGCCL~\cite{li2023sgccl} constructs a user-user graph and an item-item graph and performs edge/feature dropout to augment them. SimGCL~\cite{yu2022graph} presents experimental evidence highlighting the decisive influence of distribution consistency on recommendation performance. This study underscores the significance of distribution-based improvements over traditional dropout-based graph enhancements. It demonstrates that contrastive learning can enhance model generalization by rendering the learned representation distribution more uniform. In a parallel vein, XSimGCL~\cite{yu2023xsimgcl} and RocSE~\cite{ye2023towards} views by adding uniform noises to node representations, without using data augmentation. LightGCL~\cite{cai2023lightgcl} proposes a singular value decomposition-based graph augmentation strategy to effectively distill global collaborative signals.

Notably, NCL~\cite{lin2022improving} delves into uncovering potential relationships among nodes by characterizing high-order potential relations as enriched neighborhoods of nodes. Tailored contrastive learning methods are devised for different types of neighborhoods, enhancing the model's capacity to capture intricate node interactions. RGCL~\cite{shuai2022review} introduces review-based contrastive learning, selecting corresponding review representations as positive samples. This innovative strategy effectively harnesses the semantic richness encapsulated within reviews, contributing to more informed and context-aware recommendations. MAGCL~\cite{wangmulti} designs a multi-aspect representation learning module to decouple multi-faceted reviews and construct multi-aspect graphs to learn high-order decoupled representations. In this way, MAGCL considers multiple semantic factors existing in reviews to represent fine-grained user preferences and item features explicitly.

In summary, contrastive learning has emerged as a promising avenue for addressing data sparsity and distribution issues within recommender systems. The application of these techniques, as demonstrated by SGL, SimGCL, NCL, and RGCL, showcases the potential of contrastive learning to augment recommendation models and enrich the user experience.

\subsection{Hypergraph Learning for Recommendation}
Inspired by the inherent capability of hypergraphs to transcend the limitations of fixed and heuristic graph structures in message passing~\cite{gao2020hypergraph,zhou2006learning,bu2010music,yang2019revisiting}, recent advancements in the field of recommender systems have seen the incorporation of hypergraph connections to enhance relation learning. One noteworthy model, DualHGCN~\cite{xue2021multiplex}, adeptly transforms multiplex bipartite networks into two sets of homogeneous hypergraphs. Leveraging spectral hypergraph convolutional operators, DualHGCN employs a comprehensive approach, employing intra- and inter-message passing strategies to facilitate the exchange of information within and across different domains.

Another innovative approach, MHCN~\cite{yu2021self}, delves into social-based hypergraph convolutional networks for recommendations, exploiting various high-order user relations within a multi-channel framework. Additionally, DHCF~\cite{ji2020dual} ventures into modeling high-order correlations among users and items for broader recommendation purposes. Nevertheless, these hypergraph-based recommendation methods necessitate the manual crafting of rules to construct heuristic hypergraph structures, which remain static throughout the training process.

In response to this limitation, researchers have begun to explore dynamic approaches for learning hypergraph structures, where global collaborative relationships are modeled during the training phase. For instance, DHSL~\cite{zhang2018dynamic} employs labels as spatial feature vectors to aid in clustering hypergraph nodes, while DHGNN~\cite{jiang2019dynamic} utilizes the K-NN method and K-means clustering technique to update hypergraph structures, accounting for both local and global features. Furthermore, HCCF~\cite{xia2022hypergraph} and DHLCF~\cite{wei2022dynamic} posit that collaborative signals and topological information suffice for determining hyperedge assignments, and they employ a multi-layer perceptron (MLP) to compute these assignments. In our present work, we introduce a novel approach wherein we dynamically learn a hypergraph for each rating subgraph, enabling the capture of global collaborative relationships at a more granular level among nodes.

\begin{table}[htbp]
\centering

\caption{The summary of the main used notations.}

\label{table_notations}
\setlength{\tabcolsep}{2mm}{}

\begin{tabular}{c|c}
     \hline
     Notation & Definition \\
     \hline
     $\mathbf{U}, \mathbf{V}$  & Two types of node sets (users and items)\\
     $M$, $N$  & The number of users and items\\
     $\mathbf{R}$  & The entire user-item rating matrix \\
     $\mathcal{R}$  & The set of ratings \\
     $G$  & The multiplex user-item bipartite graph \\
     $\mathbf{E}$ & The entire learnable parameter matrix \\
     $K$ & The number of hyperedges\\
     $x^r$ & The local embeddings of nodes within the $r$-th rating subgraph\\
     $\mathcal{N}^r(i)$  & The neighbors of node $i$ in $r$-th rating subgraph\\
     $\mathbf{H}_u^r,\mathbf{H}_i^r $ & The adaptive hypergraph of nodes on the $r$-th rating subgraph\\
     $h_u^r,h_i^r$ &The global embeddings of users and items in the $r$-th rating subgraph\\
     $z,\Gamma$ & The local and global embeddings after cross-view aggregation\\
     $e$  & The final representation of nodes \\
     $d$  & The dimension of embedding \\
     $\mathbf{W},\mathbf{q}$  & The learnable weight matrix \\
     $\mathcal{N}_r$  &The two ratings that are adjacent to the rating $r$\\

     \hline
\end{tabular}
\end{table}
\section{Problem Definition}

We let $\mathbf{U}=\{u_1,u_2,\ldots,u_M\}$ denote the set of users and let $\mathbf{V}=\{v_1,v_2,\ldots,v_N\}$ denote the set of items, where $M$ and $N$ denote the number of users and items, respectively. We typically use a binary matrix $\mathbf{R}$ to store a historical rating from a user to an item and $\mathcal{R}$ is the set of ratings $(\eg,\mathcal{R}=\{1,2,3,4,5\})$, where $r_{uv}$ indicates a historical rating from a user $u$ to an item $v$ while $r_{u,v}=0$ means that item $v$ is unexposed to user $u$. Here we represent interaction data as a user-item bipartite graph $G$, and there are only interactions between users and items. The task of matrix completion is to predict missing ratings in the bipartite graph.

As previously discussed, it is noteworthy to highlight that contemporary developments in the realm of GNNs have predominantly seen the emergence of matrix completion techniques within the context of a bipartite user-item graph. This trend underscores the adoption of GNNs to tackle the challenging matrix completion task. In essence, this paradigm shift capitalizes on the inherent power of GNNs to capture complex relationships and dependencies within the user-item interaction data, thereby offering a promising avenue for advancing recommender systems and collaborative filtering approaches. Key notations used in the paper are summarized in Table~\ref{table_notations}.

\section{Methodology}

\begin{figure*}[t]
\centering

\includegraphics[scale=0.33]{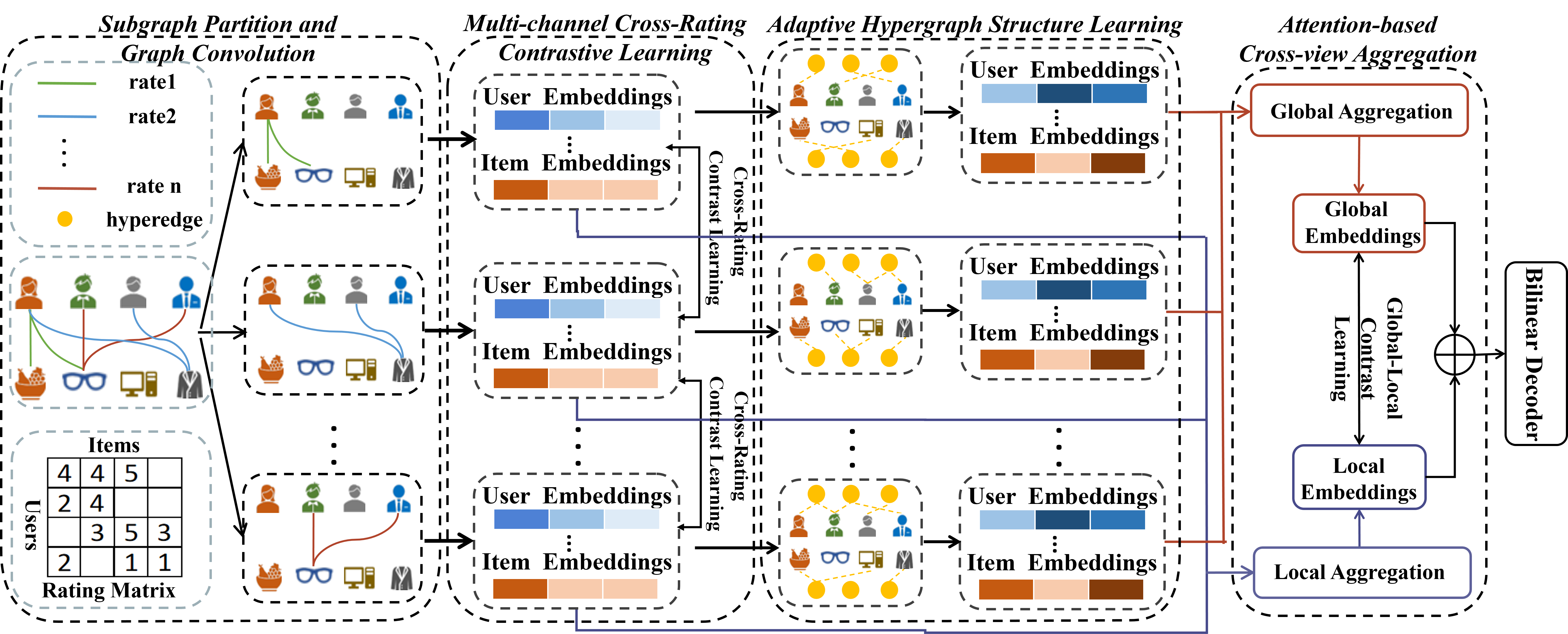} 

\caption{The outline of MHCL's overall architecture.}
\label{fig:model} 
\end{figure*}

In this section, we provide a comprehensive overview of the proposed MHCL, as illustrated in Figure~\ref{fig:model}. MHCL comprises four key components: \textbf{(1) Subgraph Partition and Graph Convolution:} MHCL begins by partitioning the bipartite user-item graph into subgraphs based on different ratings. Graph convolutions are applied to these subgraphs to capture collaborative signals and extract local topological features for users and items across various rating contexts. \textbf{(2) Multi-Channel Cross-Rating Contrastive Learning:} To enhance interaction between adjacent rating subgraphs, MHCL employs a multi-channel cross-rating contrastive learning module. This module promotes information exchange between rating channels, addressing data sparsity and mitigating the long-tail distribution problem. \textbf{(3) Adaptive Hypergraph Structure Learning:} Using local embeddings from the subgraphs, MHCL integrates an adaptive hypergraph structure learning module. This module constructs hypergraphs by identifying high-order dependencies among nodes of the same type (e.g., users with users, items with items), capturing global topological features unique to each rating. This process helps mitigate over-smoothing and enhances the model’s ability to learn higher-order correlations. \textbf{(4) Attention-Based Cross-View Aggregation:} MHCL aggregates local and global representations through an attention-based cross-view aggregation module. By maximizing the consistency between embeddings from diverse views, it leverages global-local contrastive learning techniques to obtain robust final representations.

Through these four components, MHCL delivers a comprehensive solution for matrix completion and recommendation tasks, leveraging collaborative signals, topological features, and higher-order dependencies to enhance the accuracy and efficiency of our model. 

\subsection{Subgraph Partition and Graph Convolution}
Inspired by previous models~\cite{shen2021inductive, zhang2022geometric, zhang2019inductive}, we adopt the initialization method from IMC-GAE~\cite{shen2021inductive}. Let $M$ and $N$ represent the number of users and items, respectively, and $\mathcal{|R|}$ be the number of rating categories. The initial node embeddings are parameterized by a learnable matrix $\mathbf{E} \in \mathbb{R}^{\mathcal{|R|} \times (M+N+3)\times d}$. Each node's initial features consist of three indices: identical index $\mathbf{id}$, role-aware index $\mathbf{ra}$, and one-hot index $\mathbf{oh}$. We set the identical index to 0, the user's role-aware index to 1, the item's to 2, and the one-hot index ranges from 3 to ($M$+$N$+3). Using these initial features, we derive the node's identical, role-aware, and one-hot embeddings from $\mathbf{E}$. Finally, we concatenate the three embeddings as follows:
\begin{equation}
x^r_0[i]=Concat(\mathbf{E}^r[\mathbf{id}[i]],\mathbf{E}^r[\mathbf{ra}[i]],\mathbf{E}^r[\mathbf{oh}[i]]),
\label{eq1}
\end{equation}
where $\mathbf{E}^r\in\mathbb{R}^{(M+N+3)\times d}$ denotes $r$-th trainable parameter matrix in $\mathbf{E}\in\mathbb{R}^{\mathcal{|R|} \times (M+N+3)\times d}$, and $x^r_0[i]$ denotes the initializing embedding of the $i$-th node in the $r$-th rating subgraph. The node embeddings are the input of the message-passing layer.

Then, we systematically partition the graph into distinct rating subgraphs based on the different types of rating edges. For example, with five rating types, we create five rating subgraphs, each containing node embeddings specific to its rating category. MHCL then operates independently on these subgraphs, applying graph convolutions to derive node embeddings within each subgraph. Unlike conventional GCNs, MHCL forgoes the use of a learnable matrix for feature transformation, opting for an explicit propagation process, which can be expressed as follows:
\begin{equation}
x^r_{l+1}[i]=\sum\limits_{j\in\mathcal{N}^r(i)}\frac{1}{\sqrt{|\mathcal{N}^r(i)|\cdot|\mathcal{N}^r(j)|}}x^r_l[j],
\label{eq2}
\end{equation}
where $\mathcal{N}^r(i)$ represents the neighboring node set in the $r$-th subgraph, $x^r_{l+1}[i]$ denotes the embedding of the $i$-th node at layer $l+1$ in the $r$-th rating subgraph. By performing graph convolution on diverse rating subgraphs, we obtain multiple embeddings of nodes that are specific to different rating types, and these embeddings can represent the collaborative signal and the topological information of each user and item on different subgraphs.

\subsection{Multi-Channel Cross-Rating Contrastive Learning}

In contrast to traditional heterogeneous graphs that often feature multiple edge types (\eg, click, purchase, add to cart), we encounter a unique scenario in our context where different types of ratings are inherently comparable. Typically, adjacent ratings indicate that a user's preference for an item is relatively similar, leading to correspondingly similar node representations. However, the presence of imbalanced rating distributions, stemming from the long-tail distribution inherent in real datasets, adds complexity to the situation. A critical challenge arises due to our proposed MHCL of partitioning the heterogeneous graph into distinct subgraphs based on different rating types, which inadvertently results in the loss of both magnitude and order information between the various ratings when performing graph convolution.

To address this challenge, we introduce an innovative multi-channel cross-rating contrastive learning module. We treat diverse rating subgraphs as distinct (rating) channels, each offering a unique perspective on the data, and we leverage cross-view contrastive learning to capture cooperative associations. Furthermore, we introduce mutual enhancement mechanisms to align adjacent channels, leveraging their inherent correlations to strengthen representations and bolster the effectiveness of MHCL. This module extracts richer learning signals, improving the representations of less frequent ratings and mitigating rating data sparsity. Specifically, we employ the well-established InfoNCE loss~\cite{gutmann2010noise} built upon cross-rating representations of users and items. Here, we consider a user's embedding learned from the current rating subgraph and the user embedding from the next adjacent rating subgraph as positive pairs, respecting the order of ratings. The mathematical equations are presented as follows:
\begin{equation}\label{eq3}
    {\mathcal{L}_{S(k)}^\mathbf{U}=\sum_{u\in \mathbf{U}}-\log\frac{exp((x_u^{(k)}\cdot x_u^{(k+1)}/\tau))}{\sum_{u'\in \mathbf{U-u}}exp((x_{u}^{(k)}\cdot x_{u'}^{(k+1)}/\tau))},}
\end{equation}
where $x_u^{(r)}$ is the normalized output embedding of rating subgraph $r$, $\tau$ is the temperature hyper-parameter of softmax, $u'$ represents a node distinct from node $u$. 


Most real-world datasets typically use 5-point, 10-point, and 100-point Likert scales. For instance, in a 100-point scale, directly performing contrastive learning across all ratings is impractical. To tackle this, we map the task to a 10-point Likert scale, conducting contrastive learning only between embeddings from rating subgraphs that are evenly divisible by 10.

Using a 5-point Likert scale as an example, $L_{S(1)}^\mathbf{U}$ denotes the cross-rating contrastive loss between the subgraphs with ratings `1' and `2', while $L_{S(2)}^\mathbf{U}$ represents the loss between the subgraphs with ratings `2' and `3'. The overall cross-rating contrastive loss for the entire user can be expressed as follows:
\begin{equation}
\mathcal{L}_{S}^\mathbf{U} =\mathcal{L}_{S(1)}^\mathbf{U}+\mathcal{L}_{S(2)}^\mathbf{U}+\dots+\mathcal{L}_{S(\mathcal{|R|}-1)}^\mathbf{U}.
\label{eq4}
\end{equation} 
where $\mathcal{|R|}$ denotes the number of rating categories. In a similar way, the cross-ratings contrastive loss of the item side $\mathcal{L}_{S}^\mathbf{V}$ can be obtained as:
\begin{equation}
\mathcal{L}_{S}^\mathbf{V} =\mathcal{L}_{S(1)}^\mathbf{V}+\mathcal{L}_{S(2)}^\mathbf{V}+\dots+\mathcal{L}_{S(\mathcal{|R|}-1)}^\mathbf{V},
\label{eq5}
\end{equation}
and the complete cross-rating contrastive objective function is the sum of the above two losses:
\begin{equation}
\mathcal{L}_{S} = \mathcal{L}_{S}^\mathbf{U} + \mathcal{L}_{S}^\mathbf{V}.
\label{eq6}
\end{equation}

\subsection{Adaptive Hypergraph Structure Learning}
As previously discussed, the embedding derived from Eq.~\eqref{eq2} effectively encapsulates collaborative signals and topological information about users and items within distinct rating subgraphs. It is worth noting that the inherent topological similarities among nodes often mirror implicit similarities in user preferences or item features. In light of this insight, MHCL endeavors to transcend the conventional limitation inherent in bipartite graphs where edges typically link nodes of the same type, and involves the hypergraph structure learning for both users and items. Through hypergraph convolution, MHCL adaptively captures global information that transcends the boundaries of node types. Drawing inspiration from prior methodologies~\cite{wei2022dynamic, xia2022hypergraph}, we employ the MLP in the computation of hyperedge assignments for users and items within each rating subgraph. This process yields hypergraphs $\mathbf{H}^r_u$ for users and $\mathbf{H}^r_v$ for items, each reflecting the intricate relationships and dependencies that exist within the respective rating contexts:
\begin{equation}
{\mathbf{H}^r_u}=normalize(LeakyReLU({x_u^r}\mathbf{W}_u^r)), \quad {\mathbf{H}^r_v}=normalize(LeakyReLU({x_v^r}\mathbf{W}_v^r)),
\label{eq7}
\end{equation}
where $Norm(\cdot)$ denotes the normalization function, $\mathbf{W}_u^r\in\mathbb{R}^{d\times K}$ are trainable weight matrices, in which $K$ is the number of hyperedges.


Upon deriving the user hypergraph and item hypergraph associated with various ratings, we possess the means to incorporate implicit high-order correlations among nodes into the updated representations of both users and items. Drawing inspiration from conventional hypergraph convolution techniques~\cite{feng2019hypergraph, ji2020dual}, we adhere to the established framework of spectral hypergraph convolution for its balance between computational efficiency and representational power, which aligns well with our needs. The process unfolds as follows: Initially, node embeddings are amalgamated to form hyperedge representations through hyperedge convolution. Subsequently, these hyperedge embeddings are re-aggregated to create the novel node representations. The hypergraph convolution, which operates independently on each rating subgraph, is formally defined as follows:
\begin{equation}
h_u^r={\mathbf{H}}_u^r\mathbf{D}_u^{-1}{\mathbf{H}}_u^{r\mathsf{T}}{x_u^r}, \quad h_v^r={\mathbf{H}}_v^r\mathbf{D}_v^{-1}{\mathbf{H}}_v^{r\mathsf{T}}{x_v^r}.
\label{eq8}
\end{equation}

Here, $\mathbf{D}_u$ and $\mathbf{D}_v$ represent the vertex degree matrices associated with the learned user hypergraph $\mathbf{H}_u^r$ and item hypergraph $\mathbf{H}_v^r$, respectively. These matrices play a crucial role in the rescaling of the resulting embeddings.

\subsection{Attention-based Cross-View Aggregation}

As elucidated in Eq.~\eqref{eq2} and Eq.~\eqref{eq9}, each node acquires a local embedding that captures the intricacies of the local topology, and a global embedding that encapsulates higher-order dependencies within each rating subgraph. However, in the context of downstream tasks, the need often arises for a more compact node embedding. Traditional methods commonly resort to straightforward pooling operations (such as average or sum pooling) applied to embeddings from various perspectives to obtain condensed representations. However, these operations tend to overlook the significance of node representations corresponding to diverse relations.

To address this limitation, we introduce an attention-based cross-view contrastive aggregation mechanism that aims at strengthening relationships between adjacent ratings. Formally, let $\mathcal{N}_r$ represent the two ratings that are adjacent to rating $r$, and let $x^r$ denote the local embeddings of nodes within the $r$-th rating subgraph. The process of message passing between adjacent ratings is executed as follows:
\begin{equation}
z^r=x^r+\sum\limits_{r^{\prime}\in\mathcal{N}_r)}\lambda^{r^{\prime}}\cdot x^{r^{\prime}},
\label{eq9}
\end{equation}
where $\lambda^{r^{\prime}}$ is the normalized relevance of rating $r^{\prime}$ to rating $r$, and it is calculated by:
\begin{equation}
\begin{gathered}
\lambda^{r^{\prime}}=\frac{\exp\left(L e a k y R e L U\left(q^rx^{r^{\prime}}\right)\right)}{\sum_{r^{\prime}\in \mathcal{N}_r}\exp\left(L e a k y R e L U\left(q^rx^{r^{\prime}}\right)\right)}. 
\end{gathered}
\label{eq10}
\end{equation}

The vector $q^r$ is a trainable attention vector tailored to the rating $r$. It serves as a pivotal element in regulating the information flow between relation $r$ and the neighboring relations found within $\mathcal{N}_r$. Similarly, we employ the attention-based cross-view aggregation mechanism on the global embeddings in a parallel manner:
\begin{equation}
\Gamma^r=h^r+\sum\limits_{r^{\prime}\in\mathcal{N}_r}\delta^{r^{\prime}}\cdot h^{r^{\prime}},
\label{eq11}
\end{equation}
where $h^{r^{\prime}}$ signifies the global embeddings of nodes within the ${r^{\prime}}$-th rating subgraph. Similar to $\lambda^{r'}$ in Eq.~\eqref{eq10}, $\delta^{r^{\prime}}$ can be calculated according to the similar computation process. Subsequently, we consolidate the global and local embeddings of all nodes across rating subgraphs into a unified vector representation using a summation operation. Finally, we further process this intermediate output through a linear operator:
\begin{equation}
z[u/v]=tanh(\mathbf{W}_{u/v}^z\sum\limits_{r\in{\mathcal{R}}}z^r[u/v]), \quad \Gamma[u/v]=tanh(\mathbf{W}_{u/v}^{\Gamma}\sum\limits_{r\in{\mathcal{R}}}\Gamma^r[u/v]),
\label{eq12}
\end{equation}
where $\mathbf{W}_{u/v}^z$ and $\mathbf{W}_{u/v}^{\Gamma}$ denote the weight matrices in the nonlinear activation function $tanh(\cdot)$, $z[u/v]$ and $\Gamma[u/v]$ represent local and global embeddings of users/items after cross-view aggregation. The final representation of users and items, $e[u/v]$, is the combination of local embedding and global embedding of the user/item:
\begin{equation}
e[u/v]=z[u/v]+\Gamma[u/v].
\label{eq13}
\end{equation}

Subsequently, we employ contrastive learning to enhance the complementarity between local collaborative embeddings from user-item interaction graphs and global embeddings from hypergraphs linking users and items. In this framework, different views of the same user or item form positive pairs, while divergent views form negative pairs, providing auxiliary supervision from both local and global spaces. This is formalized with a global-local contrastive loss for user representations, based on the InfoNCE framework, as follows:
\begin{equation}
\mathcal{L}_P^\mathbf{U}=\sum_{u\in \mathbf{U}}-\log\frac{exp((z[u]\cdot \Gamma[u]/\gamma))}{\sum_{u'\in \mathbf{U}-u}exp((z[u]\cdot \Gamma[u']/\gamma))}, \quad \mathcal{L}_P^\mathbf{V}=\sum_{v\in \mathbf{V}}-\log\frac{exp((z[v]\cdot \Gamma[v]/\gamma))}{\sum_{v'\in \mathbf{V}-v}exp((z[v]\cdot \Gamma[v']/\gamma))}.
\label{eq14}
\end{equation}

The final global-local contrastive loss is the sum of the user objective and the item objective:
\begin{equation}
\mathcal{L}_P=\mathcal{L}_P^\mathbf{U} + \mathcal{L}_P^\mathbf{V}.
\label{eq15}
\end{equation}

\subsection{Bilinear Decoder and Model Optimization}
In line with prior work~\cite{berg2017graph, shen2021inductive}, we adopt a bilinear decoder to reconstruct edges within the user-item graph, while considering different ratings as distinct classes. We denote the reconstructed rating matrix as $\check{M}$, which effectively captures the relationships between users and items. The decoder functions by bilinearizing the potential rating levels, generating a probability distribution, and subsequently applying a softmax function:
\begin{equation}
p(\check{\mathbf{R}}_{i j}=r)=\frac{e[i]^{\mathsf{T}}\mathbf{Q}_{r}e[j]}{\sum_{r^{\prime}=1}^{\mathcal{R}}e[i]^{\mathsf{T}}\mathbf{Q}_{r^{\prime}}e[j]},
\label{eq16}
\end{equation}
where $\mathbf{Q}_{r}$ is the learnable parameter matrix of rating $r$. The predicted rating matrix is calculated as:
\begin{equation}
\check{\mathbf{R}}=\sum_{r\in\mathcal{R}}r \cdot p\left(\check{\mathbf{R}}_{ij}=r\right).
\label{eq17}
\end{equation}

We use a bilinear decoder to frame rating prediction as a classification task, adopting Cross-Entropy (CE) loss over Mean Squared Error (MSE). While MSE is common, it often converges slowly and risks suboptimal solutions due to gradual weight updates and non-convex optimization challenges~\cite{qi2020mean,jin2023implicit}.

However, traditional CE loss overlooks the sequential and comparable nature of ratings. For example, if the true rating is 3, CE treats predictions like (0.1, 0.15, 0.3, 0.25, 0.2) and (0.05, 0.2, 0.3, 0.35, 0.1) equally, even though the latter is more accurate. To address this, we enhance CE by framing the task as multi-class classification, assigning 1 to the correct rating and a smaller value, $l_{close}$, to adjacent ratings as follows:
\begin{equation}
\mathcal{L}_{main}=\frac{1}{|(i,j)|\Omega_{i,j}=1|}\sum\limits_{(i,j)|\Omega_{i,j}=1}\mathbf{bCE}(r[i,j],\hat{r}[i,j]),
\label{eq18}
\end{equation}
where we denote the true rating label as $r[i,j]$ and the predicted rating as $\hat{r}[i,j]$ for the pair $(i,j)$. Then we introduce $\alpha$ and $\beta$ to optimize the contrastive learning tasks. This results in the final loss function, defined as follows:
\begin{equation}
\mathcal{L}=\mathcal{L}_{main}+\alpha\mathcal{L}_S+\beta\mathcal{L}_P+\lambda\mathcal{L}_{NRR}.
\label{eq19}
\end{equation}

Here, we introduce $\mathcal{L}_{NRR}$ as a regularization method inspired by prior models~\cite{berg2017graph, shen2021inductive}. This regularization method is designed to account for both magnitude and order information, which aims to foster similarity in the representation of each node in the rating subgraph that is adjacent to each other.

\subsection{Time Complexity Analysis}

The proposed MHCL consists of four key components: subgraph partition and graph convolution module, multi-channel cross-rating contrastive learning module, adaptive hypergraph structure learning module, and attention-based cross-view aggregation module. 
For the subgraph partition and graph convolution module, the time complexity is $O((M+N+3)\times \mathcal{|R|} \times L \times d)$, where $\mathcal{|R|}$ represents the number of rating categories in the user-item entire matrix, $M$ and $N$ represent the number of users and items, $L$ represents the number of GCN message propagation layer. The time complexity of multi-channel cross-rating contrastive learning module is $O((M+N)\times d^2 \times \mathcal{|R|} + (M+N)\times d \times \mathcal{|R|})$. The time complexity of the adaptive hypergraph structure learning module is $O((M+N)^2\times d)$. For the attention-based cross-view aggregation module, the time complexity is $O(4(\mathcal{|R|}-1)^2\times d)$. 
Conclusively, the total time complexity of our MHCL is $O((M+N)^2 \times d + (M+N) \times \mathcal{|R|} \times d \times (d + L + 1) + 4(\mathcal{|R|}-1)^2 \times d + 3\mathcal{|R|} \times L \times d)$.


\section{Experiment}
In this section, we evaluate the performance of our proposed \model through extensive experiments and answer the following questions: 
\begin{itemize}[leftmargin=*]
    \item \textbf{(RQ1)} How does \model effectively perform in the matrix completion task? 
    \item  \textbf{(RQ2)} How does \model perform compared to rating and review-based recommendations after matrix completion?
    \item  \textbf{(RQ3)} How does \model perform on large-scale datasets?
    \item  \textbf{(RQ4)} What are the effects of different modules in \model on performance?
    \item \textbf{(RQ5)} How does \model mitigate long-tail distribution and data sparsity problems?
    \item \textbf{(RQ6)} How do different hyperparameter settings affect the performance?
    \item \textbf{(RQ7)} How does \model perform in terms of efficiency compared to baselines?
    \item \textbf{(RQ8)} What insights can be obtained from case studies of the proposed MHCL?
\end{itemize}

\subsection{Experimental Settings}
\subsubsection{\textbf{Datasets}}
To assess the efficacy of \model, we conduct matrix completion and recommendation experiments on eight publicly available datasets, including Yelp~\cite{seo2017interpretable}, Amazon~\cite{ni2019justifying}, MovieLens-100K (ML-100K for short)~\cite{miller2003movielens}, MovieLens-1M (ML-1M for short)~\cite{miller2003movielens}, YahooMusic~\cite{dror2012yahoo}, Douban~\cite{miller2003movielens}, CDs and Vinyl (CD for short)~\cite{wang2023multi}, and Alibaba-iFashion (Alibaba for short)~\cite{chen2019pog}. The summary of dataset statistics is presented in Table~\ref{table_data}. 
For YahooMusic dataset, we utilize preprocessed subsets of this dataset as provided by~\cite{monti2017geometric}. Amazon and Yelp datasets are distinct in that they include not only user-item ratings but also detailed review information. These datasets encompass four primary data components: users, items, ratings, and review documents. We focus on the `Digital Music' domain within the Amazon dataset, specifically the Amazon-5core dataset in our experiments, where `5-core' indicates that each user or item has a minimum of five reviews. Yelp dataset originates from the Yelp Business Rating Prediction Challenge 2013 dataset, which comprises restaurant reviews from the Phoenix, AZ metropolitan area. Following the preprocessing steps outlined in~\cite{shuai2022review}, we establish a 5-core dataset from the raw Yelp data. Alibaba is a sparse dataset, and we randomly sample 300k users and use all their interactions with fashion outfits.

\begin{table*}[t]
\centering
\setlength{\tabcolsep}{5pt}  
\renewcommand{\arraystretch}{1.2}  

\caption{Statistical summaries of datasets.}

\label{table1}
\begin{tabular}{c|c|c|c|c|c|c|c|c|c|c}
\toprule
\multicolumn{2}{c|}{\multirow{2}{*}{Dataset}} &\multirow{2}{*}{\#Users}&\multirow{2}{*}{\#Items} &\multirow{2}{*}{\#Ratings} &\multirow{2}{*}{density(\%)} & \multicolumn{5}{c}{Ratings\%}\\
\cline{7-11}
\multicolumn{2}{c|}{} & & & & & 1/(1-20) &  2/(21-40) & 3/(41-60) & 4/(61-80) & 5/(81-100)\\
\midrule
\multicolumn{2}{c|}{Yelp} &8,423  &3,742	&88,647	&0.281	&5.88
&9.41	&15.29	&37.13 &32.09	\\ 
\multicolumn{2}{c|}{Amazon} &5,541  &3,568	&64,706	&0.330	&4.31 &4.65	&10.49	&25.55 &54.98	\\ 
\multicolumn{2}{c|}{ML-100K} &943  &1,682	&100,000	&6.30	&6.11 &11.37	&27.14	&34.17 &21.2	\\ 
\multicolumn{2}{c|}{ML-1M} & 6,040  & 3,706	& 1,000,209	& 4.47 & 5.61 & 10.75	& 26.12	& 34.89 & 22.63	\\ 
\multicolumn{2}{c|}{YahooMusic} & 3,000 & 3,000 & 5,335	& 0.06	& 1.04 & 11.84	& 19.15	& 19.03 & 48.94	\\ 
\multicolumn{2}{c|}{Douban} &3,000  &3,000	&136,891	&1.52	&5.87 & 11.25	& 25.71	& 30.23 & 26.94	\\
\multicolumn{2}{c|}{CD} & 75,258 & 64,443 & 1,097,592 & 0.023 & 5.11 & 11.25 & 23.26 & 29.77 & 30.61 \\
\multicolumn{2}{c|}{Alibaba} & 300,000 & 81,614 & 1,607,813 & 0.007 & 4.43 & 9.84 & 27.08 & 30.50 & 28.15 \\
\bottomrule
\end{tabular}
\label{table_data}

\end{table*}



\subsubsection{\textbf{Baselines}}
To demonstrate the effectiveness, we compare our proposed MHCL with the following four categories of methods: (1) traditional methods: \textbf{GRALS}~\cite{rao2015collaborative},  \textbf{NNMF}~\cite{dziugaite2015neural}; (2) GAE-based methods: \textbf{GC-MC}~\cite{berg2017graph}, \textbf{IMC-GAE}~\cite{shen2021inductive}; (3) inductive methods: \textbf{PinSage}~\cite{ying2018graph}, \textbf{IGMC}~\cite{zhang2019inductive}, \textbf{GIMC}~\cite{zhang2022geometric}, \textbf{IDCF-NN}~\cite{wu2021towards} and \textbf{IDCF-GC}~\cite{wu2021towards}; (4) rating-based methods: \textbf{SGL}~\cite{wu2021self}, and \textbf{VGCL}~\cite{yang2023generative}. Besides, we also introduce: (5) rating and review-based recommendation methods: \textbf{SSG}~\cite{gao2020set}, \textbf{RGCL}~\cite{shuai2022review},\textbf{MAGCL}~\cite{wang2023multi},\textbf{DGCLR}~\cite{ren2022disentangled}, and \textbf{ReHCL}~\cite{wang2024enhanced} for comparison to evaluate the recommendation performance of \model.

\noindent
\textbf{Traditional Method:}
\begin{itemize}[leftmargin=*]
    \item \textbf{GRALS}~\cite{rao2015collaborative} is a graph-regularized matrix completion algorithm, which has formulated and derived an exceptionally efficient alternating minimization scheme based on conjugate gradients. The scheme excels in solving optimizations involving over 55 million observations.
    \item \textbf{NNMF}~\cite{dziugaite2015neural} is a classical matrix factorization model which uses the user-item ratings (or interactions) only as the target value of its objective function. It explores the possibility of substituting the inner product with a learned function, concurrently with the learning process of latent feature vectors.
\end{itemize}
\textbf{GAE-based Methods:}
\begin{itemize}[leftmargin=*]
    \item \textbf{GC-MC}~\cite{berg2017graph} regards rating prediction as link prediction on the user-item bipartite graph and adopts Graph Autoencoder (GAE) to encode user and item embeddings and proposes a graph auto-encoder framework that can utilize differentiable message passing on the bipartite interaction graph.
    \item \textbf{IMC-GAE}~\cite{shen2021inductive} is an inductive GAE-based matrix completion method that leverages GAE to acquire personalized user or item representations for tailored recommendations while also capturing local graph patterns for inductive matrix completion.
\end{itemize}
\textbf{Inductive Methods:}
\begin{itemize}[leftmargin=*]
    \item \textbf{PinSage}~\cite{ying2018graph} combines a highly effective random walk and graph convolution to represent node embeddings that incorporate both node features and graph structure. It also introduces an innovative training strategy, which progressively incorporates more challenging training examples to enhance the model's robustness and convergence. 
    \item \textbf{IGMC}~\cite{zhang2019inductive} employs a GNN-based model that is exclusively trained on 1-hop subgraphs centered around (user, item) pairs derived from the rating matrix. It exhibits inductive capabilities, learning local graph patterns to generalize to new ones for inductive matrix completion.
    \item \textbf{GIMC}~\cite{zhang2022geometric} stands as the pioneer in defining continuous explicit feedback prediction within a non-Euclidean space by introducing hyperbolic geometry and a unified message-passing scheme to reflect node interaction and represent edge embeddings for improving inductive matrix completion for rating prediction.
    \item \textbf{IDCF-NN}~\cite{wu2021towards} is an inductive collaborative filtering method that adopts the traditional matrix factorization method to factorize the rating matrix of a set of key users to obtain meta-latent factors.
    \item \textbf{IDCF-GC}~\cite{wu2021towards} is an inductive collaborative filtering method, which is an attention-based structured learning method that estimates the hidden relationship from query to key users and learns to use meta-latents to inductively compute the embedding of query users through neural information transfer.
\end{itemize}
\textbf{Rating-based Methods:}
\begin{itemize}[leftmargin=*]
    \item \textbf{SGL}~\cite{wu2021self} enhances GCN-based recommendation by adding a self-supervised task. It generates multiple node views using node dropout, edge dropout, and random walk, maximizing agreement between views of the same node. Implemented on LightGCN, SGL improves recommendation accuracy, especially for long-tail items, and increases robustness to noise.
    \item \textbf{VGCL}~\cite{yang2023generative} estimates a Gaussian distribution for each node, generating contrastive views through sampling. These views reconstruct the input graph without distortion, and node-specific variance adjusts contrastive loss. A twofold contrastive learning approach ensures consistency at both the node and cluster levels.
\end{itemize}
\noindent
\textbf{Rating and Review-based Methods:}
\begin{itemize}[leftmargin=*]
    \item \textbf{SSG}~\cite{gao2020set} introduces a multi-view approach to enhance existing single-view methods, incorporating two additional perspectives: sequence and graph, for leveraging reviews. Specifically, the three-way encoder architecture captures long-term (set), short-term (sequence), and collaborative (graph) user-item features for recommendations.
    \item \textbf{RGCL}~\cite{shuai2022review} is a novel graph-based contrastive learning framework that firstly creates a review-aware user-item graph with edge features enriched by review content, and then introduces two additional contrastive learning tasks to provide self-supervised signals.
    \item \textbf{MAGCL}~\cite{wang2023multi} is a review-based recommendation method that captures fine-grained semantic information in user reviews while considering the intrinsic correlation between ratings and reviews, and learns higher-order structure-aware and self-discriminative node embeddings through multi-view representation learning, graph contrastive learning, and multi-task learning mechanisms.
    \item \textbf{DGCLR}~\cite{ren2022disentangled} models user-item interactions through latent factors derived from reviews and user-item graphs. A factorized message-passing mechanism learns disentangled representations, while an attention mechanism combines predictions from multiple factors. Factor-wise contrastive learning mitigates sparsity and improves factor-specific interaction modeling.
    \item \textbf{ReHCL}~\cite{wang2024enhanced} constructs topic and semantic graphs to mine review relations and employs cross-view contrastive learning to enhance node representations. A neighbor-based positive sampling captures similarities between views, reducing noise. Additionally, cross-modal contrastive learning aligns rating and review representations. These modes form a hierarchical contrastive learning framework to improve recommendation performance.
\end{itemize}

It is worth noting that certain methods within these groups require specific preprocessing steps before model training. For instance, RGCL employs BERT-Whitening~\cite{su2021whitening} to convert textual comments into vector representations, while both IGMC and GIMC mandate the extraction of subgraphs for all edges within the dataset to operate effectively. For most of the baselines, we split our training, validation, and test sets in the ratio of 8:1:1, following the same settings in recent studies~\cite{zhang2022geometric,shen2021inductive}.

\subsubsection{Evaluation Metrics and Experimental Settings}
We use MSE and MAE metrics for evaluating the matrix completion task, and MRR@10 (M@10 for short) and NDCG@10 (N@10 for short) for the recommendation task. 

We employ the Adam optimizer for the entirety of model training, initializing all trainable parameters using the Xavier method. We systematically explore several key hyperparameters: The number of layers is varied within the range $\{1, 2, ..., 5\}$. The hyperparameters $\alpha$ and $\beta$ in a range of $\{0.001,0.005,0.01,0.05,0.1\}$, $\lambda$ is searched in $\{$$1e^{-7}$, $1e^{-6}$, $1e^{-5}$, $1e^{-4}$$\}$. In the embedding layer, we consider different vector sizes, selecting from \{30, 60, ..., 1200\}. The number of hyperedges is searched in $\{8, 16, 32, 128, 256, 512\}$. Additionally, we implement early stopping to mitigate the risk of overfitting. The decay ratio, denoted as $\theta$, is set to 0.5 to further fine-tune training dynamics. We use early stopping with a 10-epoch patience, halting training if validation performance does not improve. The complete model implementation is carried out using the Deep Graph Library~\cite{wang2019deep} and PyTorch, harnessing the computational power of an Intel Xeon Gold 5320T CPU (2.3GHz) with two NVIDIA RTX 3090 GPUs.

Specifically, in the recommendation section, we utilize the following loss function to evaluate the model performance:
\begin{equation}
    \mathcal{L}= \sum\limits_{(u,{v_ + },{v_ - }) \in \mathcal{O}} { - \ln (Sigmoid(\hat{y}_{u,{v_ + }} - \hat{y}_{u,{v_ - }}))}  + \lambda ||\Theta |{|^2}.
\end{equation}
where $\hat{y}_{u,{v_ +}} = e_u[i]^\mathsf{T}e_{v_+}[i]$, $(u,{v_ + }) \in \mathcal{O}_+, (u,{v_ - }) \in \mathcal{O}_-$, $\mathcal{O}_+$ and $\mathcal{O}_-$ respectively denote the observed existent interaction and unobserved interaction, and $\mathcal{O}=\mathcal{O}_+ + \mathcal{O}_-$.

\subsection{Matrix Completion (RQ1)}
As shown in Table~\ref{tab.mc_baselines}, MHCL consistently outperforms all baselines across diverse datasets, with the most notable +9.04\% MSE improvement on ML-100K. MHCL excels without needing extra side information or review data, making it simpler than models requiring complex preprocessing. It also surpasses review-based models on Amazon and Yelp datasets, avoiding the significant time and memory costs associated with creating individual subgraphs for each rating edge, especially in large datasets. In the other four datasets without review information (ML-100K, ML-1M, YahooMusic, and Douban), compared to the suboptimal methods IMC-GAE and GIMC, MHCL demonstrates substantial improvements, achieving a maximum MAE performance improvement of +4.34\% on YahooMusic dataset. The performance highlights MHCL's superiority in mitigating data sparsity and long-tail distribution issues.
\begin{table*}[t!]
\begin{center}

\caption{Matrix completion performance comparison of the proposed model and different baselines. The best results are in bold and the second-best are underlined. Marker * indicates the results are statistically significant (t-test with p-value < 0.01).}

\label{tab.mc_baselines}
\setlength{\tabcolsep}{0.6mm}{}
\begin{tabular}{c|cc|cc|cc|cc|cc|cc}
\toprule
\multirow{2}{*}{Method} & \multicolumn{2}{c|}{Amazon} & \multicolumn{2}{c|}{Yelp}  & \multicolumn{2}{c|}{ML-100K}  & \multicolumn{2}{c|}{ML-1M} & \multicolumn{2}{c|}{YahooMusic} & \multicolumn{2}{c}{Douban} \\
& MSE & MAE & MSE & MAE & MSE & MAE & MSE & MAE & MSE & MAE & MSE & MAE \\
\midrule
GRALS & 0.8297 & 0.6879 & 1.1553 & 0.8462 & 0.8930 
& 0.7481 & 0.6905 & 0.6511 & 1444 & 51.26 & 0.8625 & 0.7304 \\ 
NNMF & 0.8316 & 0.6960 	& 1.1607 & 0.8530      & 0.8154 & 0.7116    & 0.7106 & 0.6663 	& 466.56 & 18.43 & 0.8658 & 0.7355 \\
\midrule
GC-MC   & 0.8090 & 0.6788  & 1.1508 & 0.8429     & 0.8190 & 0.7164    & 0.6922 & 0.6520     & 420.25 & 17.01  & 0.8596 & 0.7159 \\ 
IMC-GAE   & 0.7921 & 0.6635  & 1.1418 & \underline{0.8389}  & 0.8046 & 0.7068         & \underline{0.6872} & \underline{0.6486}     & 349.69 & 14.77 & 0.8533 & 0.7092 \\ 
\midrule
PinSage & 0.8447 & 0.7032 & 1.1715 & 0.8615 & 0.9044 & 0.7522 & 0.8208 & 0.6773 & 524.41 & 20.33 & 0.8636 & 0.7641 \\
IGMC & 0.7903 & 0.6616 & 1.1481 & 0.8452 & 0.8190 & 0.7132 & 0.7344 & 0.6773 & 364.81 & 15.41 & 0.8491 & 0.7235 \\ 
GIMC & 0.7889 & 0.6597 & 1.1435 & 0.8422 & \underline{0.8010} & \underline{0.7034} & 0.7276 & 0.6724 & \underline{338.56} & \underline{14.52} & \underline{0.8401} & \underline{0.7092} \\ 
IDCF-NN  & 0.7823 & 0.6536    & 1.1415 & 0.8403     & 0.8085 & 0.7059    & 0.6901 & 0.6532   & 342.67 & 15.02  & 0.8419 & 0.7110 \\ 
IDCF-GC   & 0.7791 & 0.6527     & 1.1402 & 0.8391     & 0.8030 & 0.7035    & 0.6889 & 0.6501    & 340.01 & 14.75  & 0.8402 & 0.7098 \\
\midrule
SSG & 0.8218 & 0.6780 & 1.1613 & 0.8529 & $\backslash$ & $\backslash$ & $\backslash$ & $\backslash$ & $\backslash$ & $\backslash$ & $\backslash$ & $\backslash$
\\ 
RGCL & 0.7735 & \underline{0.6524} & \underline{1.1397} & 0.8394 & $\backslash$ & $\backslash$ & $\backslash$ & $\backslash$ & $\backslash$ & $\backslash$ & $\backslash$ & $\backslash$
\\ 
MAGCL & \underline{0.7730} & 0.6525 & 1.1398 & 0.8390 & $\backslash$ & $\backslash$ & $\backslash$ & $\backslash$ & $\backslash$ & $\backslash$ & $\backslash$ & $\backslash$
\\ 
\midrule
\textbf{MHCL} & \textbf{0.7610*} & \textbf{0.6350*} & \textbf{1.1289*} & \textbf{0.8255*} & \textbf{0.7790*} & \textbf{0.6907*} & \textbf{0.6790*} & \textbf{0.6361*} & \textbf{334.89*} & \textbf{13.89*} & \textbf{0.8298*} & \textbf{0.7019*}
\\ 
\bottomrule
\end{tabular}
\end{center}

\end{table*}

Traditional matrix completion methods, such as GRALS and NNMF, perform well on small datasets but struggle with scalability, while GNN-based models excel in this area. Although GC-MC and IMC-GAE provide fast runtimes, they converge more slowly and deliver slightly inferior performance compared to MHCL. Inductive models like IGMC and GIMC achieve competitive results but require substantial time investment, with their performance on certain datasets not matching that of MHCL. Models such as PinSage, IDCF-NN, and IDCF-GC effectively capture local patterns but fail to model higher-order dependencies and cross-rating features, which diminishes their accuracy in ordinal matrix completion tasks. Review-based models like SSG and RGCL leverage user comments to enhance embeddings but cannot function without review data. In contrast, MHCL dynamically learns hypergraph structures for each rating subgraph, capturing both local and global dependencies, and integrates these representations through cross-rating fusion. Notably, MHCL does not rely on review data yet consistently outperforms review-based methods on datasets like Amazon and Yelp, showcasing its versatility. Moreover, MHCL offers significant advantages in time complexity, making it a scalable and efficient solution for large-scale matrix completion tasks.

\begin{table*}[t!]
\begin{center}
\caption{Recommendation performance comparison of the proposed model and different baselines. The best results are in bold and the second-best are underlined. Marker * indicates the results are statistically significant (t-test with p-value < 0.01).}

\label{tab.r_baselines}
\setlength{\tabcolsep}{0.5mm}{}
\begin{tabular}{c|cc|cc|cc|cc|cc|cc}
\toprule
\multirow{2}{*}{Method} & \multicolumn{2}{c|}{Amazon} & \multicolumn{2}{c|}{Yelp}  & \multicolumn{2}{c|}{ML-100K}  & \multicolumn{2}{c|}{ML-1M} & \multicolumn{2}{c|}{YahooMusic} & \multicolumn{2}{c}{Douban} \\
& M@10 & N@10 & M@10 & N@10 & M@10 & N@10 & M@10 & N@10 & M@10 & N@10 & M@10 & N@10 \\
\midrule
GRALS & 0.2529 & 0.3301 & 0.2705 & 0.3364 & 0.2920 & 0.3495 & 0.3921 & 0.4133 & 0.2920 & 0.3524 & 0.3147 & 0.3655 \\ 
NNMF & 0.2497 & 0.3299 & 0.2701 & 0.3360 & 0.2911 & 0.3487 & 0.3907 & 0.4415 & 0.2909 & 0.3510 & 0.3134 & 0.3642 \\
\midrule
GC-MC & 0.2555 & 0.3310 & 0.2713 & 0.3377 & 0.2928 & 0.3501 & 0.3948 & 0.4453 & 0.2942 & 0.3542 & 0.3166 & 0.3670 \\ 
IMC-GAE & 0.2557 & 0.3314 & 0.2721 & 0.3387 & 0.2939 & 0.3511 & 0.3950 & 0.4454 & 0.2945 & 0.3547 & 0.3168 & 0.3671 \\ 
\midrule
PinSage & 0.2565 & 0.3330 & 0.2740 & 0.3411 & 0.2961 & 0.3529 & 0.3958 & 0.4466 & 0.2955 & 0.3555 & 0.3180 & 0.3682 \\
IGMC & 0.2603 & 0.3366 & 0.2768 & 0.3430 & 0.2973 & 0.3550 & 0.3988 & 0.4493 & 0.2980 & 0.3583 & 0.3201 & 0.3705 \\ 
GIMC & 0.2637 & 0.3379 & 0.2787 & 0.3448 & 0.3002 & 0.3570 & 0.4009 & 0.4527 & 0.2996 & 0.3607 & 0.3208 & 0.3717 \\ 
IDCF-NN  & 0.2791 & 0.3409 & 0.2802 & 0.3465 & 0.3011 & 0.3585 & 0.4028 & 0.4569 & 0.3018 & 0.3625 & 0.3220 & 0.3730 \\ 
IDCF-GC  & 0.2825 & 0.3417 & 0.2810 & 0.3479 & \underline{0.3013} & \underline{0.3597} & \underline{0.4035} & \underline{0.4588} & \underline{0.3027} & \underline{0.3645} & \underline{0.3229} & \underline{0.3735} \\
\midrule
SGL & 0.2915 & 0.3472 & 0.2929 & 0.3498 & $\backslash$ & $\backslash$ & $\backslash$ & $\backslash$ & $\backslash$ & $\backslash$ & $\backslash$ & $\backslash$
\\
VGCL & 0.2941 & 0.3496 & 0.2960 & 0.3562 & $\backslash$ & $\backslash$ & $\backslash$ & $\backslash$ & $\backslash$ & $\backslash$ & $\backslash$ & $\backslash$
\\
SSG & 0.2807 & 0.3398 & 0.2795 & 0.3472 & $\backslash$ & $\backslash$ & $\backslash$ & $\backslash$ & $\backslash$ & $\backslash$ & $\backslash$ & $\backslash$
\\ 
RGCL & 0.2845 & 0.3477 & 0.2834 & 0.3501 & $\backslash$ & $\backslash$ & $\backslash$ & $\backslash$ & $\backslash$ & $\backslash$ & $\backslash$ & $\backslash$
\\
MAGCL & 0.2948 & 0.3569 & 0.2899 & 0.3597 & $\backslash$ & $\backslash$ & $\backslash$ & $\backslash$ & $\backslash$ & $\backslash$ & $\backslash$ & $\backslash$
\\
DGCLR & 0.2943 & 0.3502 & 0.2956 & 0.3555 & $\backslash$ & $\backslash$ & $\backslash$ & $\backslash$ & $\backslash$ & $\backslash$ & $\backslash$ & $\backslash$
\\
ReHCL & \underline{0.3050} & \underline{0.3611} & \underline{0.3020} & \underline{0.3627} & $\backslash$ & $\backslash$ & $\backslash$ & $\backslash$ & $\backslash$ & $\backslash$ & $\backslash$ & $\backslash$
\\
\midrule
\textbf{MHCL} & \textbf{0.3153*} & \textbf{0.3768*} & \textbf{0.3130*} & \textbf{0.3815*} & \textbf{0.3317*} & \textbf{0.3968*} & \textbf{0.4214*} & \textbf{0.4972*} & \textbf{0.3233*} & \textbf{0.3926*} & \textbf{0.3471*} & \textbf{0.4035*} 
\\
\bottomrule
\end{tabular}
\end{center}

\end{table*}

\subsection{Recommendation (RQ2)}
Next, we evaluate the recommendation performance by comparing MHCL with the latest recommendation baselines which consider user rating matrices. Expect baselines in Table~\ref{tab.mc_baselines}, we add the review and rating-based recommendation baselines (SGL, DGCLR, VGCL, ReHCL). Experimental results are reported in Table~\ref{tab.r_baselines}, MHCL consistently outperforms all advanced baselines across datasets, with the most notable +10.31\% N@10 improvement on ML-100K and average +6.65\% improvement on M@10 and N@10 across all datasets. 
The performance highlights MHCL's superiority in recommendation tasks.

In recommendation tasks, many matrix-completion methods integrate user rating matrices into node representations to effectively capture user preferences and enhance overall recommendation performance. For instance, on Amazon and Yelp, which include user reviews, the current state-of-the-art method, ReHCL, leverages both the rating matrix and the textual reviews provided by users. In contrast, our method, MHCL, achieves an average performance improvement of +4.14\% over ReHCL without incorporating any review data, relying solely on the rating matrix. This outcome highlights the effectiveness and necessity of our matrix completion algorithm design, demonstrating that a well-structured rating matrix can be sufficient for improving recommendation quality. Furthermore, when analyzing four datasets that do not contain reviews, we observe an average performance improvement of +10.20\% in M@10 and N@10 metrics on ML-100K dataset compared to IDCF-GC, the leading matrix completion method among the baselines. This finding suggests that the rating matrices generated through MHCL not only capture more nuanced user preference information but also maintain their effectiveness across diverse datasets. Overall, MHCL's ability to deliver superior performance without relying on review data underscores its potential as a robust solution for matrix completion in recommender systems.

\subsection{Performance on large-scale datasets (RQ3)}
\begin{table}[t]
\centering
\caption{Performance comparison of models on large-scale graph datasets.}
\setlength{\tabcolsep}{3mm}{}	
\begin{tabular}{c|cccc|cccc}
\toprule
\multirow{4}{*}{Method} & \multicolumn{4}{c|}{CD} & \multicolumn{4}{c}{Alibaba}  \\
\cmidrule(lr){2-5} \cmidrule(l){6-9}
& \multicolumn{2}{c|}{Matrix Completion} & \multicolumn{2}{c|}{Recommendation} & \multicolumn{2}{c|}{Matrix Completion} & \multicolumn{2}{c}{Recommendation} \\
\cmidrule(lr){2-5} \cmidrule(l){6-9}
& MSE & \multicolumn{1}{c|}{MAE} & M@10 & N@10 & MSE & \multicolumn{1}{c|}{MAE} & M@10 & N@10 \\
\midrule
GRALS & 0.8558 & \multicolumn{1}{c|}{0.6980} & 0.3509 & 0.4326 & 2.0459 & \multicolumn{1}{c|}{1.4434} & 0.1049 & 0.0460 \\
NNMF & 0.8640 & \multicolumn{1}{c|}{0.7077} & 0.3515 & 0.4340 & 2.0547 & \multicolumn{1}{c|}{1.4550} & 0.1054 & 0.0466 \\
\midrule
GC-MC & 0.8404 & \multicolumn{1}{c|}{0.6828} & 0.3533 & 0.4351 & 2.0305 & \multicolumn{1}{c|}{1.4327} & 0.1070 & 0.0485 \\
IMC-GAE & 0.8234 & \multicolumn{1}{c|}{0.6796} & 0.3546 & 0.4368 & 2.0140 & \multicolumn{1}{c|}{1.4156} & 0.1069 & 0.0486 \\
\midrule
PinSage & 0.8726 & \multicolumn{1}{c|}{0.7151} & 0.3555 & 0.4382 & 2.0613 & \multicolumn{1}{c|}{1.4620} & 0.1081 & 0.0498 \\
IGMC & 0.8215 & \multicolumn{1}{c|}{0.6767} & 0.3559 & 0.4388 & OOM. & \multicolumn{1}{c|}{OOM.} & OOM. & OOM. \\
GIMC & 0.8210 & \multicolumn{1}{c|}{0.6759} & 0.3579 & 0.4403 & OOM. & \multicolumn{1}{c|}{OOM.} & OOM. & OOM. \\
IDCF-NN & 0.8201 & \multicolumn{1}{c|}{0.6751} & 0.3597 & 0.4415 & 2.0101 & \multicolumn{1}{c|}{1.4119} & 0.1095 & 0.0514 \\
IDCF-GC & 0.8193 & \multicolumn{1}{c|}{0.6725} & 0.3607 & 0.4450 & 2.0089 & \multicolumn{1}{c|}{1.4109} & 0.1101 & 0.0517 \\
\midrule
SGL & 0.8151 & \multicolumn{1}{c|}{0.6691} & 0.3685 & 0.4452 & 2.0079 & \multicolumn{1}{c|}{1.4090} & 0.1118 & 0.0525 \\
VGCL & 0.8121 & \multicolumn{1}{c|}{0.6654} & 0.4030 & 0.4786 & \underline{2.0066} & \multicolumn{1}{c|}{\underline{1.4075}} & \underline{0.1131} & \underline{0.0540} \\
SSG & 0.8458 & \multicolumn{1}{c|}{0.6853} & 0.3665 & 0.4415 & $\backslash$ & \multicolumn{1}{c|}{$\backslash$} & $\backslash$ & $\backslash$ \\
RGCL & 0.8180 & \multicolumn{1}{c|}{0.6705} & 0.3618 & 0.4479 & $\backslash$ & \multicolumn{1}{c|}{$\backslash$} & $\backslash$ & $\backslash$ \\
MAGCL & \underline{0.8099} & \multicolumn{1}{c|}{\underline{0.6609}} & 0.4110 & 0.4860 & $\backslash$ & \multicolumn{1}{c|}{$\backslash$} & $\backslash$ & $\backslash$ \\
DGCLR & 0.8132 & \multicolumn{1}{c|}{0.6670} & 0.3676 & 0.4520 & $\backslash$ & \multicolumn{1}{c|}{$\backslash$} & $\backslash$ & $\backslash$ \\
ReHCL & 0.8102 & \multicolumn{1}{c|}{0.6629} & \underline{0.4115} & \underline{0.4863} & $\backslash$ & \multicolumn{1}{c|}{$\backslash$} & $\backslash$ & $\backslash$ \\
\midrule
{\textbf{MHCL}} & \textbf{0.7655} & \multicolumn{1}{c|}{\textbf{0.6259}} & \textbf{0.4471} & \textbf{0.5280} & \textbf{1.9789} & \multicolumn{1}{c|}{\textbf{1.3858}} & \textbf{0.1239} & \textbf{0.0595} \\
\bottomrule
\end{tabular}
\label{tab:l_performance}
\end{table}

To evaluate \model on large-scale datasets CD and Alibaba, we compare it with SOTA methods on both matrix completion and recommendation tasks. Notice that review-based methods do not directly apply to the Alibaba dataset, as they do not contain review data. To facilitate a fair comparison, SGL, DGCLR, VGCL, and ReHCL, originally designed for recommendation, have been adapted to the matrix completion task using a unified loss function and evaluation metrics. Results in Table \ref{tab:l_performance} highlight the clear superiority of MHCL over all baselines. Specifically, in the matrix completion task, MHCL achieves up to +5.48\% improvement on two datasets, demonstrating its enhanced ability to model complex relationships between users and items while maintaining model robustness. In the recommendation task, MHCL significantly outperforms rating and review-based recommendation methods, marking a substantial average +8.61\% and +13.81\% improvement on CD and Alibaba datasets, respectively. These results underscore MHCL’s effectiveness in capturing fine-grained dependencies while scaling efficiently, making it capable of being applied to large-scale graphs.

\subsection{Ablation Study (RQ4)}\label{ablation}
\begin{figure*}[t]
\centering

\subfigure[MSE for matrix completion.]{
	    \label{abla_mc}
		\includegraphics[scale=0.35]{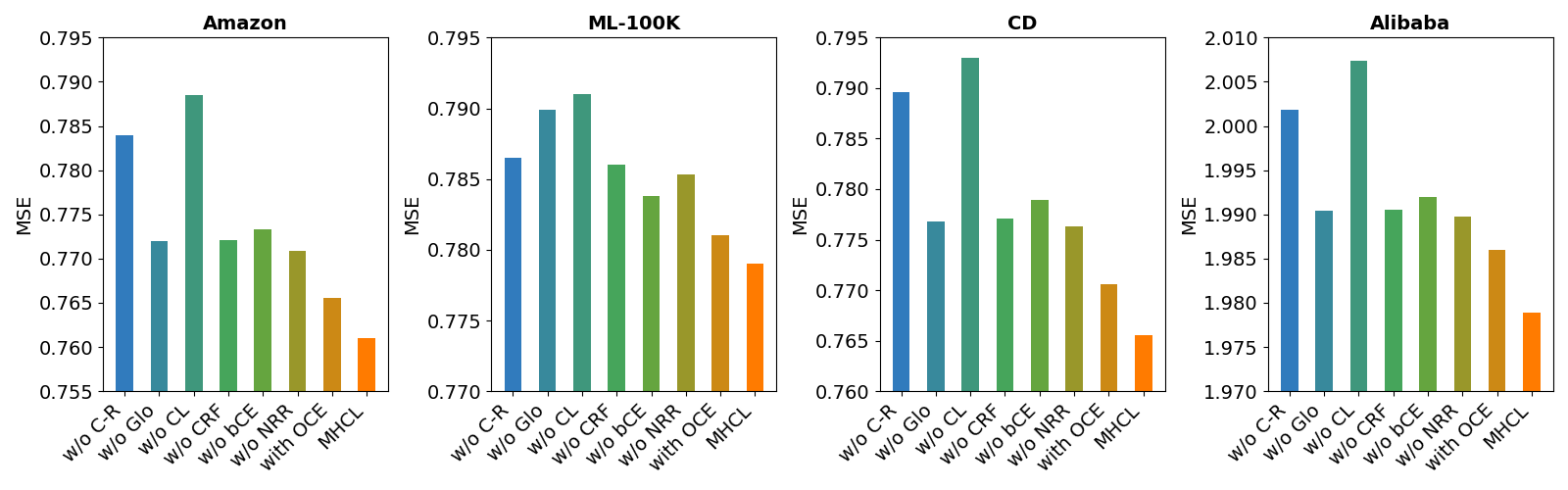} 
	}
 \subfigure[NDCG@10 for recommendation.]{
        \label{abla_r}
        \includegraphics[scale=0.35]{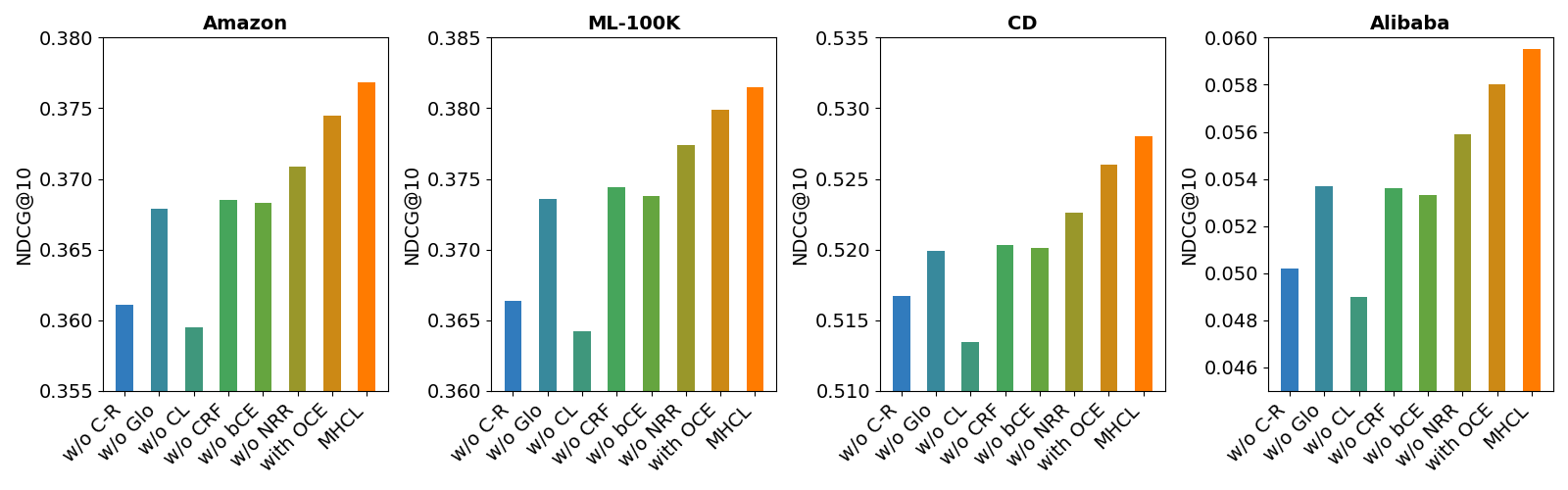}
 }
\caption{Performance comparison of different variants.}
\label{fig:ablation} 
\end{figure*}

To evaluate the effectiveness of each component of our model, we have further conducted an ablation study on different MHCL variations. We report the results of the ablation study on four datasets in Figure~\ref{fig:ablation}, with MSE metric in the matrix completion task and NDCG@10 metric in the recommendation task. Specifically, we generate the following variants:
\begin{itemize}[leftmargin=*]
\item \textbf{w/o C-R} - which removes the multi-channel cross-rating contrastive learning and corresponds to the $\alpha \mathcal{L}_S$ part of the final loss function. 
\item \textbf{w/o Glo} - which removes the adaptive hypergraph structure learning module and only learns the local node embeddings.
This variant corresponds to the $\beta\mathcal{L}_P$ part in the final loss function. 
\item \textbf{w/o CL} - which removes all contrastive learning mechanisms.
\item \textbf{w/o CRF} - which removes the cross-rating representation fusing module.
\item \textbf{w/o bCE} - which uses the traditional cross-entropy loss instead of the balanced CE loss that captures the nuances of ratings. 
\item \textbf{w/o NRR} - which removes the $\lambda \mathcal{L}_{NRR}$ in the final loss function.
\item \textbf{with OCE} - which uses the ordinal cross-entropy loss (OCE)~\cite{shi2023deep} instead of the bCE loss that captures the nuances of ratings.
\end{itemize}

Our results yield several key insights. In the matrix completion task: First, removing the cross-ratings representation fusion (CRF) module (\textbf{w/o CRF}) leads to performance degradation across all datasets, underscoring its critical importance in capturing complex user interactions. Similarly, omitting the multi-channel cross-ratings module and global hypergraph learning (\textbf{w/o C-R} and \textbf{w/o Glo}) also results in reduced performance. Notably, the cross-ratings module significantly enhances results for Amazon, where the rating distribution is imbalanced, while hypergraph learning offers particular benefits for denser datasets such as ML-100K and ML-1M. The variant without contrastive learning (\textbf{w/o CL}) demonstrates the poorest performance, highlighting the crucial role of contrastive learning in enriching node representations and improving model robustness. Additionally, removing the balanced cross-entropy loss (\textbf{w/o bCE}) adversely affects performance, further confirming its effectiveness in capturing correlations between closely-rated scores. Furthermore, our analysis indicates that the cross-ratings module (\textbf{C-R}) has the most substantial impact on the loss function, whereas node relation regularization (\textbf{NRR}) exhibits the least effect, aligning with our parameter sensitivity experimental results. Finally, while the ordinal cross-entropy loss (\textbf{with OCE}) performs similarly to the balanced cross-entropy loss, the latter is preferred due to its slight performance edge and superior computational efficiency, making it a more practical choice for implementation.

In the recommendation task: First, removing all contrastive learning mechanisms (w/o CL) leads to a significant drop in performance, highlighting the crucial role of contrastive learning in enhancing recommendation accuracy. Second, the absence of the multi-channel cross-rating contrastive learning module (w/o C-R) also substantially degrades model performance, especially on long-tail and sparse datasets such as Amazon, indicating its effectiveness in addressing data sparsity and imbalanced distributions. Third, eliminating the global hypergraph structure learning module (w/o Glo) has the greatest impact on dense datasets, suggesting its importance in capturing higher-order dependencies and improving model generalization. Fourth, the removal of the cross-rating fusion module (w/o CRF) results in a consistent decrease in recommendation performance across all datasets, confirming the benefit of information fusion in modeling complex user behaviors. Fifth, the design of the loss function is equally vital; replacing the original balanced cross-entropy with either the standard or ordinal variant (w/o bCE, with OCE) results in noticeable performance degradation. Finally, the ablation of the node relation regularization (w/o NRR) shows minimal impact on overall performance.

\subsection{Impact of imbalanced data distribution (RQ5)}
\label{data_sparsity}

\subsubsection{\textbf{Impact of Imbalanced User Interactions}}
We evaluate MHCL's performance across users with varying interaction sparsity by dividing them into three groups: `Inactive' (bottom 80\% by interaction frequency), `Normal' (middle 15\%), and `Active' (top 5\%). We then compare MHCL with two leading baselines on two datasets, with results shown in Table~\ref{tab.data_dist}. MHCL consistently outperforms the baselines, with the largest gains observed among Inactive users. This improvement is driven by InfoNCE loss, which mitigates popularity bias by enhancing representation uniformity.

\subsubsection{\textbf{Impact of Imbalanced Rating Interactions}}
Experimental results according to different rating categories are shown in Table~\ref{tab.data_dist}. Notably, ratings 3, 4, and 5 dominate in both the Amazon and ML-100K datasets, while ratings 1 and 2 are significantly sparser (less than 10\% in Amazon, 20\% in ML-100K). This finding highlights significant improvements in long-tail ratings (`1' and `2') without sacrificing performance in the head classes, which underscores the effectiveness of our MHCL in reshaping embedding distributions to better capture long-tail characteristics. While most methods perform well on the majority ratings, our approach significantly outperforms others on the much sparser ratings 1 and 2, demonstrating its effectiveness in handling data sparsity and long-tail distributions.
\begin{table*}[t!]
\begin{center}
\caption{Impact of imbalanced data distribution in Amazon and ML-100K datasets.}
\label{tab.data_dist}
\setlength{\tabcolsep}{2.1mm}{}
\begin{tabular}{c|c|ccc|ccccc}
\toprule
\multirow{2}{*}{Dataset} & \multirow{2}{*}{Method} & \multicolumn{3}{c|}{Each user group} & \multicolumn{5}{c}{Each rating group} \\
& & Inactive & Normal & Active & 1-point & 2-point & 3-point & 4-point & 5-point \\
\midrule
\multirow{3}{*}{Amazon} & GIMC & 0.8910 & 0.7477 & 0.6262 & 5.3752 & 3.2765 & 1.2233 & \textbf{0.2355} & \underline{0.4482} \\ 
& RGCL & \underline{0.8856} & \underline{0.7465} & \underline{0.6238} & \underline{5.1629} & \underline{3.1658} & \underline{1.2189} & \underline{0.2364} & 0.4486 \\
& \textbf{MHCL} & \textbf{0.8751} & \textbf{0.7381} & \textbf{0.6234} & \textbf{4.6893} & \textbf{2.9268} & \textbf{1.2170} & 0.2380 & \textbf{0.4480} \\ 
\midrule
\multirow{3}{*}{ML-100K} & GIMC & 0.8275 & 0.7615 & \underline{0.7358} & \underline{2.9593} & \underline{1.4625} & \underline{0.3947} & \textbf{0.2769} & 1.1116 \\ 
& RGCL & \underline{0.8266} & \underline{0.7622} & 0.7360 & 2.9850 & 1.4698 & \textbf{0.3944} & 0.2790 & \underline{1.1005} \\
& \textbf{MHCL} & \textbf{0.8074} & \textbf{0.7484} & \textbf{0.7317} & \textbf{2.7040} & \textbf{1.3690} & 0.3986 & \underline{0.2790} & \textbf{1.0894} \\ 
\bottomrule
\end{tabular}
\end{center}

\end{table*}
    
    
\subsection{Parameter Sensitivity (RQ6)}
\begin{figure*}[t]
	\centering 
	\subfigure[MSE \wrt ~$\alpha$ on Amazon]{
	    \label{alpha_amazon}
		\includegraphics[width=3.6cm]{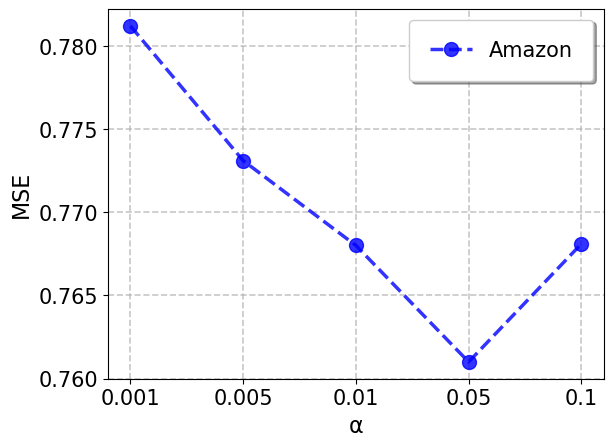} 
	}
    \hspace{-2mm}
    \subfigure[MSE \wrt ~$\beta$ on Amazon]{
	    \label{beta_amazon}
		\includegraphics[width=3.6cm]{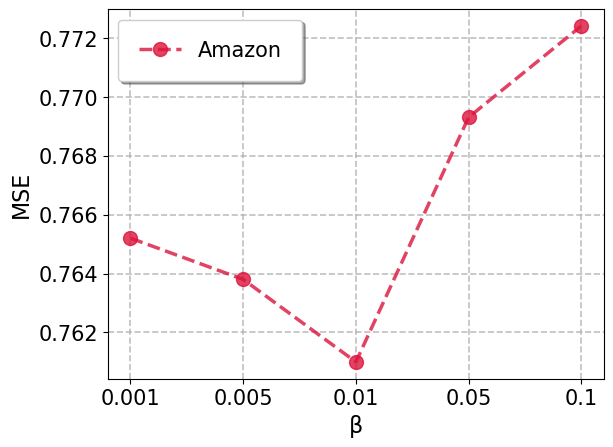} 
	}
    \hspace{-2mm}
    \subfigure[MSE \wrt ~$\lambda$ on Amazon]{
	    \label{lambda_amazon}
		\includegraphics[width=3.6cm]{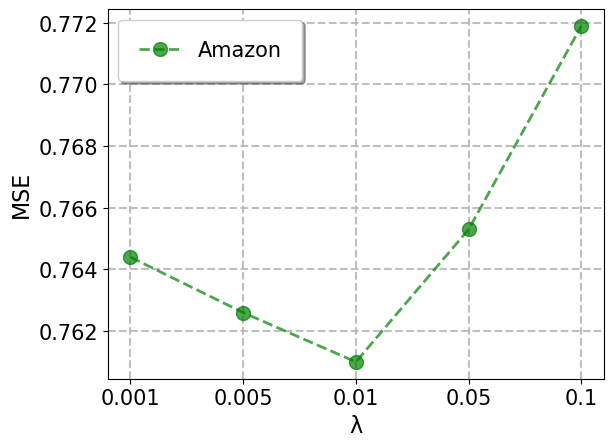} 
	}
    \hspace{-2mm}
    \subfigure[hyperedge number on Amazon ]{
	    \label{r_amazon}
		\includegraphics[width=3.6cm]{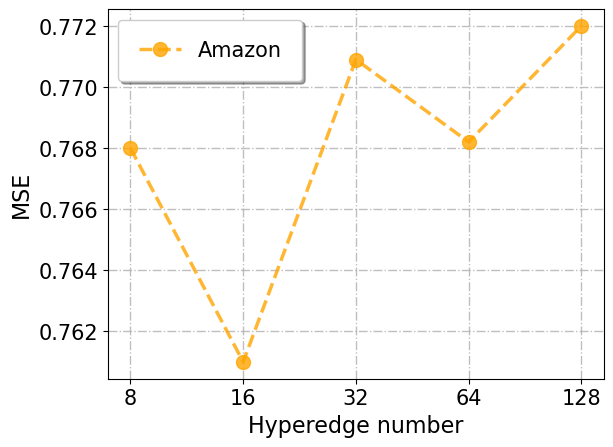} 
	}
    \subfigure[MSE \wrt ~$\alpha$ on ML-100K]{
	    \label{alpha_ml}
		\includegraphics[width=3.6cm]{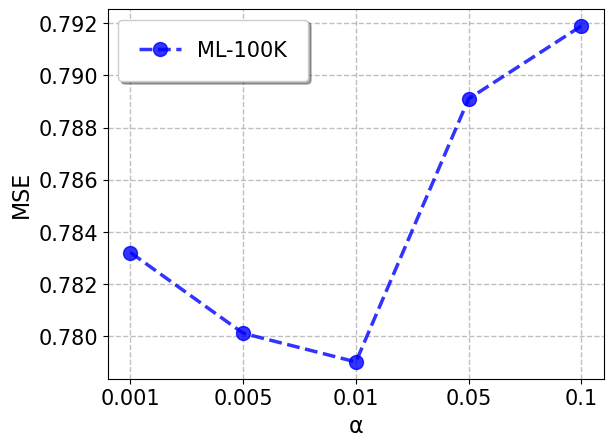}
	}
    \hspace{-2mm}
        \subfigure[MSE \wrt ~$\beta$ on ML-100K]{
	    \label{beta_ml}
		\includegraphics[width=3.6cm]{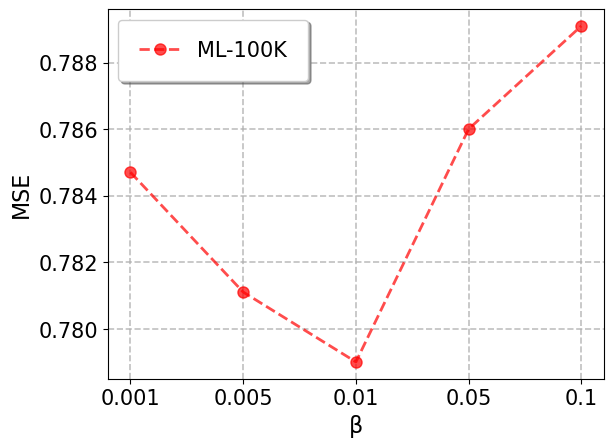}
	}
    \hspace{-2mm}
        \subfigure[MSE \wrt ~$\lambda$ on ML-100K]{
	    \label{lambda_ml}
		\includegraphics[width=3.6cm]{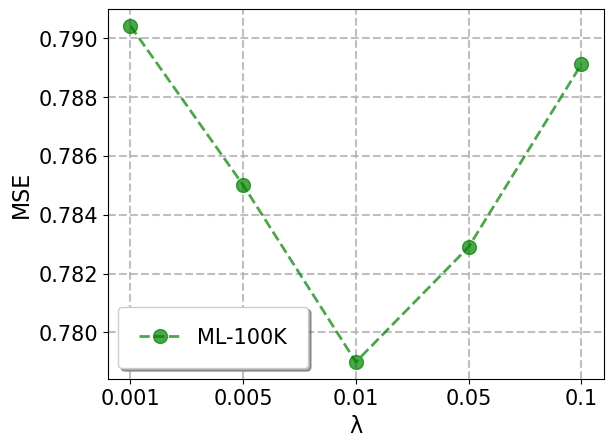}
	}
    \hspace{-2mm}
        \subfigure[hyperedge number on on ML-100K]{
	    \label{r_ml}
		\includegraphics[width=3.6cm]{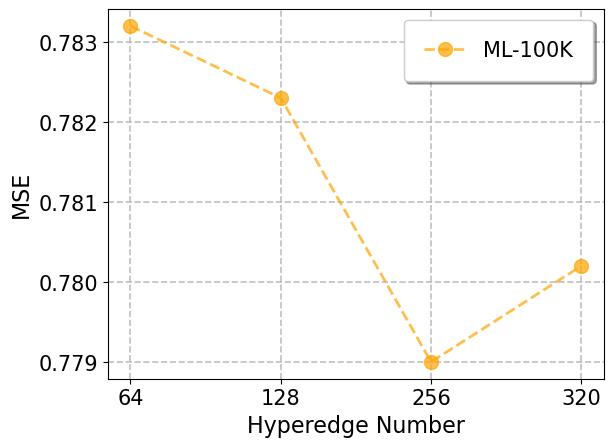} 
	}
    \caption{The parameter sensitivity experimental results.}
    \label{fig:para}
    
\end{figure*}
\subsubsection{Effect of Hyperparameters $\alpha$, $\beta$ and $\lambda$.}
We analyze the sensitivity of parameters $\alpha$, $\beta$, and $\lambda$ on contrastive learning (CL) losses by conducting parameter sensitivity experiments with $\alpha$, $\beta$, and $\lambda$ on Amazon and ML-100K datasets, using values from \{0.001, 0.005, 0.01, 0.05, 0.1\}. Results for $\alpha$, $\beta$ and $\lambda$ are shown in Figures~\ref{alpha_amazon}, \ref{alpha_ml}, \ref{beta_amazon}, \ref{beta_ml}, \ref{lambda_amazon} and \ref{lambda_ml}. 
The detailed settings of $\alpha$, $\beta$, and $\lambda$ for each dataset are as follows:
(1) Yelp: $\alpha$ = 0.05, $\beta$ = 0.01, $\lambda$ = 0.01;
(2) Amazon: $\alpha$ = 0.05, $\beta$ = 0.01, $\lambda$ = 0.01;
(3) ML-100K: $\alpha$ = 0.01, $\beta$ = 0.01, $\lambda$ = 0.01;
(4) ML-1M: $\alpha$ = 0.01, $\beta$ = 0.01, $\lambda$ = 0.01;
(5) YahooMusic: $\alpha$ = 0.01, $\beta$ = 0.01, $\lambda$ = 0.005;
(6) Douban: $\alpha$ = 0.05, $\beta$ = 0.01, $\lambda$ = 0.01;
(7) CD: $\alpha$ = 0.05, $\beta$ = 0.01, $\lambda$ = 0.01;
(8) Alibaba: $\alpha$ = 0.01, $\beta$ = 0.01, $\lambda$ = 0.01.
Increasing these parameters from 0 to 0.01 improves MHCL's performance, indicating that contrastive learning enhances node representation learning. However, excessively large values diminish the model's main loss function, negatively impacting performance. The optimal range for $\alpha$ and $\beta$ is between 0.01 and 0.05, highlighting the need to balance the main loss with contrastive learning for optimal results. Moreover, despite having fewer users and items, MHCL requires more hyperedges to capture complex high-order correlations.

\begin{figure*}[t]
	\centering 
	\subfigure[Amazon]{
	    \label{l_amazon}
		\includegraphics[width=3.6cm]{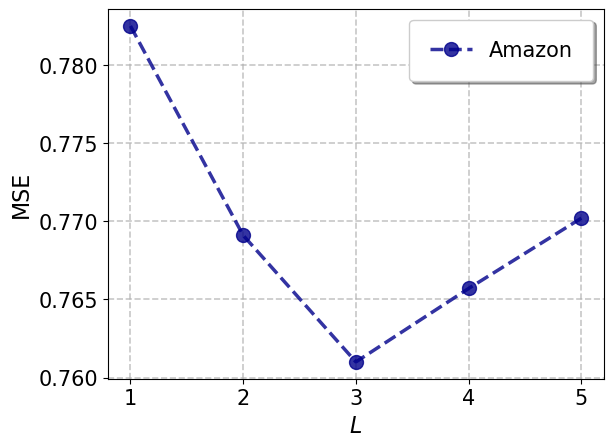} 
	}
    \hspace{-2mm}
    \subfigure[ML-100K]{
	    \label{l_ml100k}
		\includegraphics[width=3.6cm]{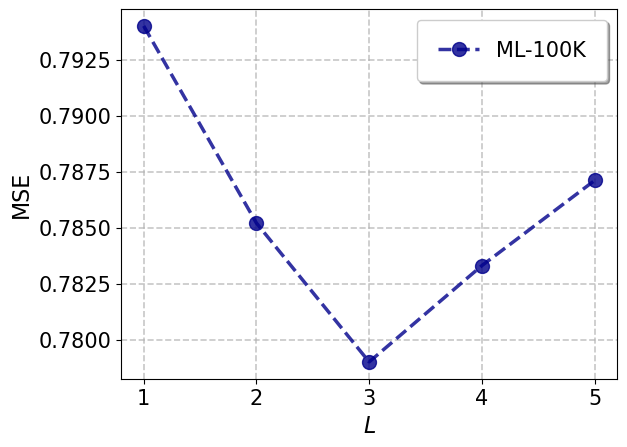} 
	}
    \hspace{-2mm}
    \subfigure[CD]{
	    \label{l_cd}
		\includegraphics[width=3.6cm]{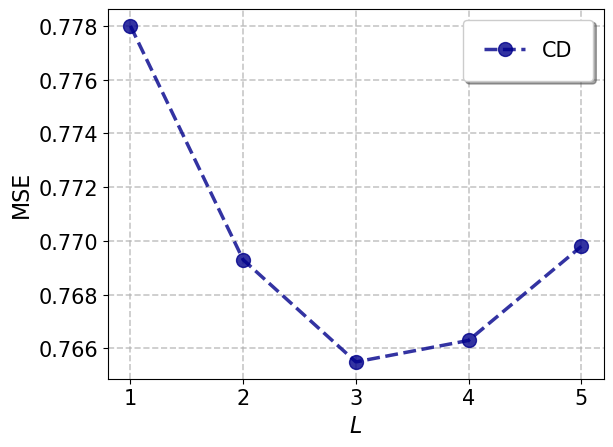} 
	}
    \hspace{-2mm}
    \subfigure[Alibaba]{
	    \label{l_alibaba}
		\includegraphics[width=3.6cm]{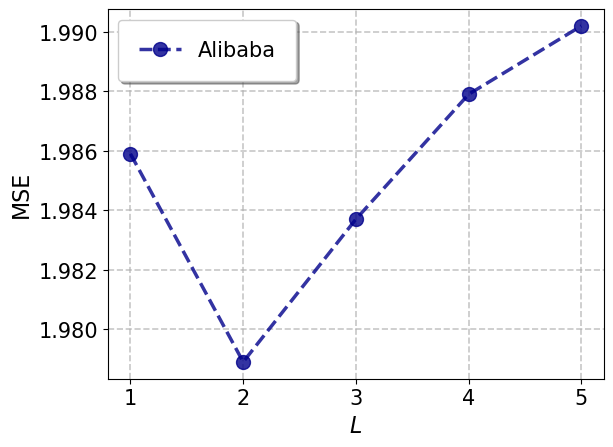} 
	}
    \caption{MSE comparison with different layers on different datasets.}
    \label{fig:layer}
\end{figure*}
\subsubsection{Effect of Hyperedge Number}
We conduct experiments by varying the hyperedge numbers, ranging from 8 to 128 on the Amazon dataset and from 64 to 320 on ML-100K dataset, while keeping all other parameters constant. The resulting insights are presented in Figure~\ref{r_amazon} and Figure~\ref{r_ml}. For the Amazon dataset, the optimal number of hyperedges appears to be 16, while on the ML-100K dataset, the model's performance reaches its peak at 256 hyperedges. Interestingly, further increasing the number of hyperedges beyond these optimal values tends to degrade the model's performance to varying degrees. This phenomenon can be attributed to the fact that an excessive number of hyperedges can introduce unnecessary noise into the representations, diluting their effectiveness. Notice that the optimal number of hyperedges is contingent upon the specific dataset's complexity. For instance, when comparing the ML-100K dataset to Amazon, despite having fewer users and items, ML-100K exhibits a denser nature. This denseness implies the existence of more intricate high-order correlations between users (items) in ML-100K, necessitating a greater number of hyperedges to adequately capture these complex relationships. Conversely, the larger user and item populations in Amazon, combined with lower density, suggest less complex user (item) dependencies, thus requiring fewer hyperedges to effectively represent these relationships.

\subsubsection{Effect of Convolution Layers.}
To investigate whether MHCL can benefit from multiple stacked layers and identify the optimal depth, we conduct experiments by varying the model's depth from 0 to 5 on Amazon and ML-100K datasets. Results illustrated in Figure~\ref{fig:layer} reveal intriguing insights. Initially, we observe that the performance of MHCL consistently improves with an increasing number of layers. However, a notable trend emerges when the number of layers reaches 3 to 4, where the model's performance starts to decline both on two datasets. This intriguing observation suggests that as we increase the number of layers, the model becomes adept at explicitly capturing high-order collaborative signals by learning the hypergraph structure across different layers. This capability effectively mitigates the common over-smoothing phenomenon associated with stacking multiple layers. Notably, \model exhibits the capacity to accommodate deeper layers compared to other GNN-based models, such as RGCL~\cite{shuai2022review} (1 layer) and IMC-GAE~\cite{shen2021inductive} (2 layers). 
This highlights MHCL's potential for capturing more intricate relationships and dependencies from dataset. 

\subsection{Efficiency Evaluation (RQ7)}
\begin{table}[t]
\begin{center}
\caption{Running time (s) for each epoch and total runtime in different datasets.}
\label{tab.time}
\setlength{\tabcolsep}{1.6mm}{}
\newcommand{\tabincell}[2]{\begin{tabular}{@{}#1@{}}#2\end{tabular}}
\small
\centering
\begin{tabular}{c|c c|c c|c c|c c|c c|c c}
\toprule
\multirow{2}{*}{\tabincell{c}{Method}} & \multicolumn{2}{c|}{ML-100K} & \multicolumn{2}{c|}{Yelp} & \multicolumn{2}{c|}{Amazon} & \multicolumn{2}{c|}{YahooMusic} & \multicolumn{2}{c|}{CD} & \multicolumn{2}{c}{Alibaba} \\ 
\cmidrule{2-13}
 & Epoch & Total & Epoch & Total & Epoch & Total & Epoch & Total & Epoch & Total & Epoch & Total \\
\midrule

GC-MC & 0.15 & 2.25 & 1.07 & 16.05 & 0.45 & 6.75 & 0.35 & 5.25 & 10.45 & 157.54 & 41.80 & 627.02 \\
IMC-GAE & 0.23 & 3.45 & 1.17 & 17.55 & 0.65 & 9.75 & 0.37 & 5.55 & 11.12 & 166.50 & 45.59 & 683.88 \\
IGMC & 706.06 & 7,060.64  & 362.22 & 3,622.20 & 267.15 & 2,671.53  & 19.33 & 193.34 & 579.30 & 5,793.03  & OOM. & OOM. \\ 
GIMC  & 599.89 & 5,998.90 & 231.16 & 2,311.64 & 143.40 & 1,434.02 & 12.75 & 127.54 & 381.31 & 3,813.11 & OOM. & OOM. \\
IDCF-NN & 0.56 & 5.62 & 2.67 & 26.71 & 1.46 & 14.58 & 0.97 & 9.69 & 29.08 & 291.02 & 116.32 & 1,163.21 \\
IDCF-GC & 0.60 & 6.04 & 2.85 & 28.53 & 1.59 & 15.91 & 1.01 & 10.14 & 30.42 & 304.21 & 124.72 & 1,247.22 \\
SSG  & 0.76 & 11.44 & 3.01 & 45.15  & 1.75 & 26.25 & 0.93 & 13.95 
& 27.99 & 419.85 & 112.01 & 1,680.15 \\ 
RGCL  & 0.46 & 6.96 & 2.41 & 36.15 & 1.14 & 17.1 & 0.79 & 11.85 & 23.67 & 355.05 & 97.05 & 1,455.71 \\
MAGCL  & 0.75 & 11.25 & 2.97 & 44.55 & 1.79 & 26.85 & 1.01 & 15.15 & 30.33 & 454.95 & 121.32 & 1,819.80 \\
MHCL & 0.45 & 4.51 & 2.42 & 24.24 & 1.13 & 11.29 & 0.75 & 7.53 & 22.52 & 225.23 & 87.83 & 878.28 \\
\bottomrule
\end{tabular}
\end{center}
\end{table}

We proceed to assess the actual running time, denoted as $t$, expended by the compared methods for a single epoch, and model total time cost which counts the total time for our method to reach convergence on the six datasets. These results are depicted in Table~\ref{tab.time}. We scrutinize the inference time of MHCL in comparison to GC-MC, IMC-GAE, IGMC, GIMC, IDCF-NN, IDCF-GC, SSG, RGCL, and MAGCL across six datasets. Notably, the two inductive models, IGMC and GIMC, incur a higher time overhead. This is primarily due to the necessity for these models to extract a local subgraph for all rating edges and subsequently relabel the nodes within each subgraph. As improved inductive approaches, IDCF-NN and IDCF-GC provide a huge improvement in time cost compared to IGMC. According to~\cite{shen2021inductive}, for a graph with $n$ nodes and $r$ ratings, GAE-based models such as GC-MC and IMC-GAE require applying the GNN $O(n)$ times to compute an embedding for each node. In contrast, IGMC and GIMC demand applying the GNN $O(r)$ times for all ratings. When the ratio of $r$ to $n$ is significantly large ($r \gg n$), IGMC exhibits worse time complexity compared to GAE-based models. Meanwhile, GIMC, by introducing edge embeddings to capture the semantic properties of different ratings rather than treating edges as propagation channels within a conventional message-passing scheme, boasts a lower time complexity than IGMC. For review-based methods such as SSG, RGCL, and MAGCL, their time complexity is significantly higher than that of our proposed MHCL due to the additional comment information they need to process, and the actual results on several datasets are slightly inferior to those of MHCL.

Note that although our MHCL consumes more total time than GC-MC and IMC-GAE, in practice our method converges faster than the above two methods, which keeps the actual total consumed time all in the same order of magnitude, and our method does not lag behind much, and MHCL achieves a better performance.

Furthermore, owing to the construction of hypergraphs and the influence of contrastive learning, MHCL demonstrates slightly higher time complexity than IMC-GAE. In particular, the incorporation of hypergraph structure learning and contrastive learning contributes to the additional computational overhead in MHCL. 




\begin{figure}
    \begin{center}
    \includegraphics[width=\textwidth]{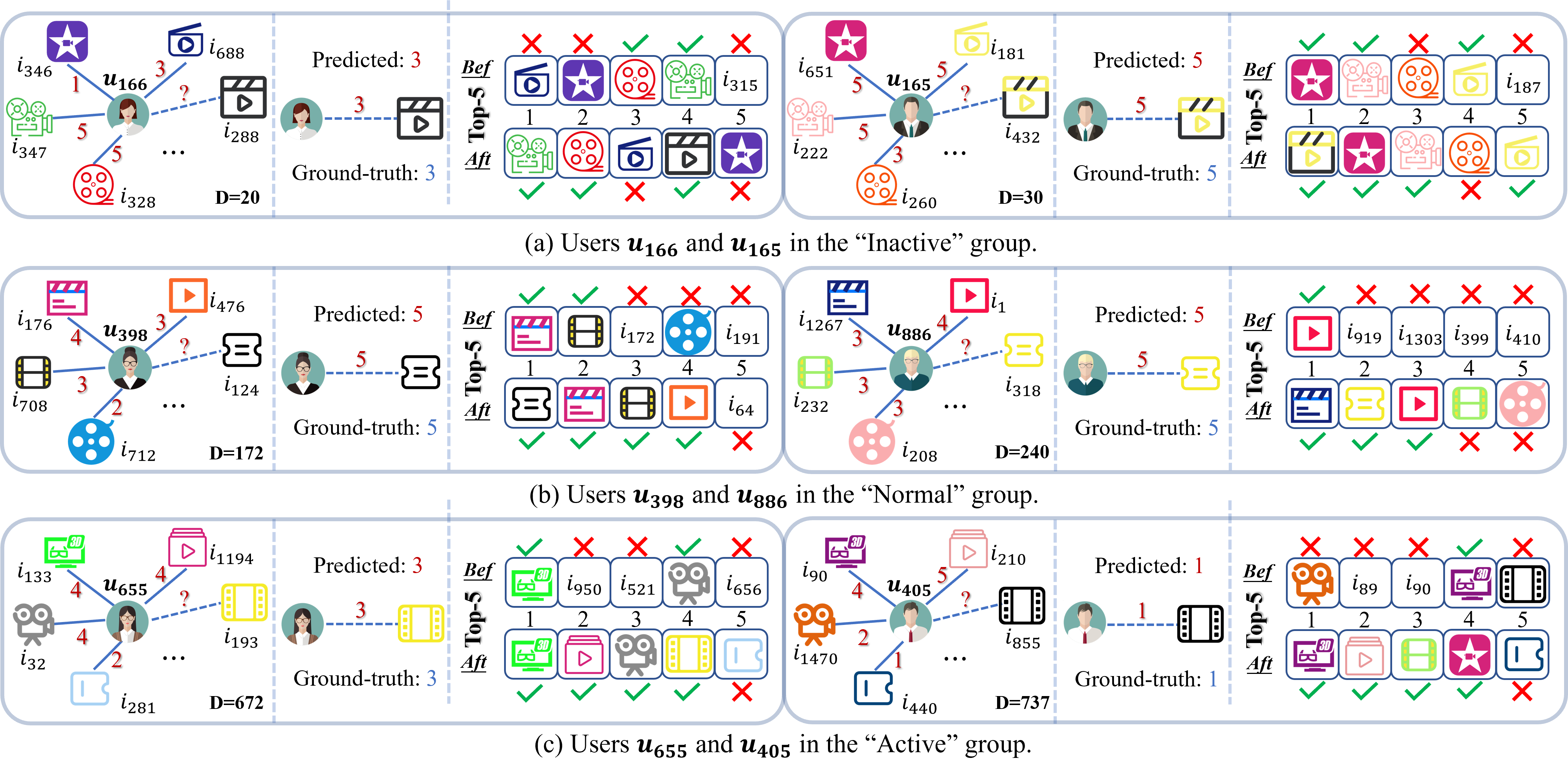}
    \caption{Case study: interaction graphs, rating matrix completion, and recommendation top-5 lists of six central nodes from `Inactive' ($u_{165}$ and $u_{166}$), `Normal' ($u_{398}$ and $u_{886}$), and `Active' ($u_{655}$ and $u_{405}$). `D' represents the degree of the central node, `\underline{\textit{Bef}}' and `\underline{\textit{Aft}}' denote recommendation \textbf{\textit{before}} and \textbf{\textit{after}} rating matrix completion.}
    \label{fig:case_study} 
    \end{center}
    \vspace{-4mm}
\end{figure}

\subsection{Case Study (RQ8)}\label{case_study}
To further illustrate the effectiveness of our \model, we conduct a detailed case study on the ML-100K dataset. As shown in Figure~\ref{fig:case_study}, we select six representative users from the `Active' (top 5\% by interaction degree), `Normal' (middle 15\%), and `Inactive' (bottom 80\%) groups based on their historical interaction frequency. For each user, we present: (1) the ego-centric user-item interaction subgraph, (2) the visualization of rating matrix completion, and (3) the comparative top-5 recommendation results before and after matrix completion.

From the visualization results, MHCL effectively leverages observed interactions to complement missing entries in the rating matrix, thereby enabling more accurate recommendations. For instance, for the `Normal' user $u_{398}$, rating matrix completion increases the number of true high-rated items in the top-5 recommendation list (\eg, $i_{124}$, $i_{176}$, $i_{476}$, and $i_{708}$). Similarly, for the `Active' user $u_{405}$, the number of correct recommendations rises from one to four after completion. This consistent trend across all six users demonstrates that integrating the completed rating matrix yields a higher hit rate and improved ranking of relevant items.
For users with abundant interactions (\eg, $u_{655}$ and $u_{405}$), MHCL fully exploits rich behavioral and structural information to guide completion and enhance recommendations. Even for users with sparse histories (\eg, $u_{166}$ and $u_{165}$), MHCL can infer missing ratings via higher-order global adaptive hypergraph structure learning and multi-channel cross-rating contrastive learning, leading to more informative and accurate top-5 recommendations. These improvements are further evidenced by the increased number of true positive matches in the recommendation lists (see green check marks in Figure~\ref{fig:case_study}).

Overall, these case study results provide both visual and quantitative evidence that MHCL achieves high-fidelity rating matrix completion and substantially boosts the accuracy and coverage of personalized recommendations for users. This strongly supports the robustness and practical value of our method in real-world scenarios.
\section{Conclusion}
This paper introduces a novel framework named MHCL for matrix completion. First, we use a hypergraph learner to construct hypergraph structures corresponding to different rating subgraphs, capturing high-order dependencies and obtaining global node embeddings through HGNN. Second, we enhance the representation of ratings by encouraging alignment between adjacent ratings. Additionally, we employ global-local contrastive learning to simultaneously characterize local and global embeddings in a joint space. Extensive experiments on five benchmark datasets demonstrate the superiority and effectiveness of our proposed MHCL.

\bibliographystyle{ACM-Reference-Format}
\bibliography{tois}


\begin{thebibliography}{84}


\ifx \showCODEN    \undefined \def \showCODEN     #1{\unskip}     \fi
\ifx \showISBNx    \undefined \def \showISBNx     #1{\unskip}     \fi
\ifx \showISBNxiii \undefined \def \showISBNxiii  #1{\unskip}     \fi
\ifx \showISSN     \undefined \def \showISSN      #1{\unskip}     \fi
\ifx \showLCCN     \undefined \def \showLCCN      #1{\unskip}     \fi
\ifx \shownote     \undefined \def \shownote      #1{#1}          \fi
\ifx \showarticletitle \undefined \def \showarticletitle #1{#1}   \fi
\ifx \showURL      \undefined \def \showURL       {\relax}        \fi
\providecommand\bibfield[2]{#2}
\providecommand\bibinfo[2]{#2}
\providecommand\natexlab[1]{#1}
\providecommand\showeprint[2][]{arXiv:#2}

\bibitem[Arrar et~al\mbox{.}(2024)]%
        {arrar2024comprehensive}
\bibfield{author}{\bibinfo{person}{Djihad Arrar}, \bibinfo{person}{Nadjet Kamel}, {and} \bibinfo{person}{Abdelaziz Lakhfif}.} \bibinfo{year}{2024}\natexlab{}.
\newblock \showarticletitle{A comprehensive survey of link prediction methods}.
\newblock \bibinfo{journal}{\emph{The journal of supercomputing}} \bibinfo{volume}{80}, \bibinfo{number}{3} (\bibinfo{year}{2024}), \bibinfo{pages}{3902--3942}.
\newblock


\bibitem[Berg et~al\mbox{.}(2017)]%
        {berg2017graph}
\bibfield{author}{\bibinfo{person}{Rianne van~den Berg}, \bibinfo{person}{Thomas~N Kipf}, {and} \bibinfo{person}{Max Welling}.} \bibinfo{year}{2017}\natexlab{}.
\newblock \showarticletitle{Graph convolutional matrix completion}.
\newblock \bibinfo{journal}{\emph{arXiv preprint arXiv:1706.02263}} (\bibinfo{year}{2017}).
\newblock


\bibitem[Bu et~al\mbox{.}(2010)]%
        {bu2010music}
\bibfield{author}{\bibinfo{person}{Jiajun Bu}, \bibinfo{person}{Shulong Tan}, \bibinfo{person}{Chun Chen}, \bibinfo{person}{Can Wang}, \bibinfo{person}{Hao Wu}, \bibinfo{person}{Lijun Zhang}, {and} \bibinfo{person}{Xiaofei He}.} \bibinfo{year}{2010}\natexlab{}.
\newblock \showarticletitle{Music recommendation by unified hypergraph: combining social media information and music content}. In \bibinfo{booktitle}{\emph{Proceedings of the 18th ACM international conference on Multimedia}}. \bibinfo{pages}{391--400}.
\newblock


\bibitem[Cai et~al\mbox{.}(2023)]%
        {cai2023lightgcl}
\bibfield{author}{\bibinfo{person}{Xuheng Cai}, \bibinfo{person}{Chao Huang}, \bibinfo{person}{Lianghao Xia}, {and} \bibinfo{person}{Xubin Ren}.} \bibinfo{year}{2023}\natexlab{}.
\newblock \showarticletitle{LightGCL: Simple Yet Effective Graph Contrastive Learning for Recommendation}.
\newblock \bibinfo{journal}{\emph{arXiv preprint arXiv:2302.08191}} (\bibinfo{year}{2023}).
\newblock


\bibitem[Candes and Recht(2012)]%
        {candes2012exact}
\bibfield{author}{\bibinfo{person}{Emmanuel Candes} {and} \bibinfo{person}{Benjamin Recht}.} \bibinfo{year}{2012}\natexlab{}.
\newblock \showarticletitle{Exact matrix completion via convex optimization}.
\newblock \bibinfo{journal}{\emph{Commun. ACM}} \bibinfo{volume}{55}, \bibinfo{number}{6} (\bibinfo{year}{2012}), \bibinfo{pages}{111--119}.
\newblock


\bibitem[Cao et~al\mbox{.}(2021)]%
        {cao2021grammatical}
\bibfield{author}{\bibinfo{person}{Hannan Cao}, \bibinfo{person}{Wenmian Yang}, {and} \bibinfo{person}{Hwee~Tou Ng}.} \bibinfo{year}{2021}\natexlab{}.
\newblock \showarticletitle{Grammatical error correction with contrastive learning in low error density domains}. In \bibinfo{booktitle}{\emph{Findings of the Association for Computational Linguistics: EMNLP 2021}}. \bibinfo{pages}{4867--4874}.
\newblock


\bibitem[Chen et~al\mbox{.}(2020)]%
        {chen2020simple}
\bibfield{author}{\bibinfo{person}{Ting Chen}, \bibinfo{person}{Simon Kornblith}, \bibinfo{person}{Mohammad Norouzi}, {and} \bibinfo{person}{Geoffrey Hinton}.} \bibinfo{year}{2020}\natexlab{}.
\newblock \showarticletitle{A simple framework for contrastive learning of visual representations}. In \bibinfo{booktitle}{\emph{International conference on machine learning}}. PMLR, \bibinfo{pages}{1597--1607}.
\newblock


\bibitem[Chen et~al\mbox{.}(2019)]%
        {chen2019pog}
\bibfield{author}{\bibinfo{person}{Wen Chen}, \bibinfo{person}{Pipei Huang}, \bibinfo{person}{Jiaming Xu}, \bibinfo{person}{Xin Guo}, \bibinfo{person}{Cheng Guo}, \bibinfo{person}{Fei Sun}, \bibinfo{person}{Chao Li}, \bibinfo{person}{Andreas Pfadler}, \bibinfo{person}{Huan Zhao}, {and} \bibinfo{person}{Binqiang Zhao}.} \bibinfo{year}{2019}\natexlab{}.
\newblock \showarticletitle{POG: personalized outfit generation for fashion recommendation at Alibaba iFashion}. In \bibinfo{booktitle}{\emph{Proceedings of the 25th ACM SIGKDD international conference on knowledge discovery \& data mining}}. \bibinfo{pages}{2662--2670}.
\newblock


\bibitem[Chen and Wang(2022)]%
        {chen2022review}
\bibfield{author}{\bibinfo{person}{Zhaoliang Chen} {and} \bibinfo{person}{Shiping Wang}.} \bibinfo{year}{2022}\natexlab{}.
\newblock \showarticletitle{A review on matrix completion for recommender systems}.
\newblock \bibinfo{journal}{\emph{Knowledge and Information Systems}} (\bibinfo{year}{2022}), \bibinfo{pages}{1--34}.
\newblock


\bibitem[Dror et~al\mbox{.}(2012)]%
        {dror2012yahoo}
\bibfield{author}{\bibinfo{person}{Gideon Dror}, \bibinfo{person}{Noam Koenigstein}, \bibinfo{person}{Yehuda Koren}, {and} \bibinfo{person}{Markus Weimer}.} \bibinfo{year}{2012}\natexlab{}.
\newblock \showarticletitle{The yahoo! music dataset and kdd-cup’11}. In \bibinfo{booktitle}{\emph{Proceedings of KDD Cup 2011}}. PMLR, \bibinfo{pages}{3--18}.
\newblock


\bibitem[Dziugaite and Roy(2015)]%
        {dziugaite2015neural}
\bibfield{author}{\bibinfo{person}{Gintare~Karolina Dziugaite} {and} \bibinfo{person}{Daniel~M Roy}.} \bibinfo{year}{2015}\natexlab{}.
\newblock \showarticletitle{Neural network matrix factorization}.
\newblock \bibinfo{journal}{\emph{arXiv preprint arXiv:1511.06443}} (\bibinfo{year}{2015}).
\newblock


\bibitem[Feng et~al\mbox{.}(2019)]%
        {feng2019hypergraph}
\bibfield{author}{\bibinfo{person}{Yifan Feng}, \bibinfo{person}{Haoxuan You}, \bibinfo{person}{Zizhao Zhang}, \bibinfo{person}{Rongrong Ji}, {and} \bibinfo{person}{Yue Gao}.} \bibinfo{year}{2019}\natexlab{}.
\newblock \showarticletitle{Hypergraph neural networks}. In \bibinfo{booktitle}{\emph{Proceedings of the AAAI conference on artificial intelligence}}, Vol.~\bibinfo{volume}{33}. \bibinfo{pages}{3558--3565}.
\newblock


\bibitem[Gao et~al\mbox{.}(2020a)]%
        {gao2020set}
\bibfield{author}{\bibinfo{person}{Jingyue Gao}, \bibinfo{person}{Yang Lin}, \bibinfo{person}{Yasha Wang}, \bibinfo{person}{Xiting Wang}, \bibinfo{person}{Zhao Yang}, \bibinfo{person}{Yuanduo He}, {and} \bibinfo{person}{Xu Chu}.} \bibinfo{year}{2020}\natexlab{a}.
\newblock \showarticletitle{Set-sequence-graph: A multi-view approach towards exploiting reviews for recommendation}. In \bibinfo{booktitle}{\emph{Proceedings of the 29th ACM International Conference on Information \& Knowledge Management}}. \bibinfo{pages}{395--404}.
\newblock


\bibitem[Gao et~al\mbox{.}(2020b)]%
        {gao2020hypergraph}
\bibfield{author}{\bibinfo{person}{Yue Gao}, \bibinfo{person}{Zizhao Zhang}, \bibinfo{person}{Haojie Lin}, \bibinfo{person}{Xibin Zhao}, \bibinfo{person}{Shaoyi Du}, {and} \bibinfo{person}{Changqing Zou}.} \bibinfo{year}{2020}\natexlab{b}.
\newblock \showarticletitle{Hypergraph learning: Methods and practices}.
\newblock \bibinfo{journal}{\emph{IEEE Transactions on Pattern Analysis and Machine Intelligence}} \bibinfo{volume}{44}, \bibinfo{number}{5} (\bibinfo{year}{2020}), \bibinfo{pages}{2548--2566}.
\newblock


\bibitem[Giorgi et~al\mbox{.}(2020)]%
        {giorgi2020declutr}
\bibfield{author}{\bibinfo{person}{John Giorgi}, \bibinfo{person}{Osvald Nitski}, \bibinfo{person}{Bo Wang}, {and} \bibinfo{person}{Gary Bader}.} \bibinfo{year}{2020}\natexlab{}.
\newblock \showarticletitle{Declutr: Deep contrastive learning for unsupervised textual representations}.
\newblock \bibinfo{journal}{\emph{arXiv preprint arXiv:2006.03659}} (\bibinfo{year}{2020}).
\newblock


\bibitem[Gutmann and Hyv{\"a}rinen(2010)]%
        {gutmann2010noise}
\bibfield{author}{\bibinfo{person}{Michael Gutmann} {and} \bibinfo{person}{Aapo Hyv{\"a}rinen}.} \bibinfo{year}{2010}\natexlab{}.
\newblock \showarticletitle{Noise-contrastive estimation: A new estimation principle for unnormalized statistical models}. In \bibinfo{booktitle}{\emph{Proceedings of the thirteenth international conference on artificial intelligence and statistics}}. JMLR Workshop and Conference Proceedings, \bibinfo{pages}{297--304}.
\newblock


\bibitem[Hamilton et~al\mbox{.}(2017)]%
        {hamilton2017inductive}
\bibfield{author}{\bibinfo{person}{Will Hamilton}, \bibinfo{person}{Zhitao Ying}, {and} \bibinfo{person}{Jure Leskovec}.} \bibinfo{year}{2017}\natexlab{}.
\newblock \showarticletitle{Inductive representation learning on large graphs}.
\newblock \bibinfo{journal}{\emph{Advances in neural information processing systems}}  \bibinfo{volume}{30} (\bibinfo{year}{2017}).
\newblock


\bibitem[He et~al\mbox{.}(2020)]%
        {he2020lightgcn}
\bibfield{author}{\bibinfo{person}{Xiangnan He}, \bibinfo{person}{Kuan Deng}, \bibinfo{person}{Xiang Wang}, \bibinfo{person}{Yan Li}, \bibinfo{person}{Yongdong Zhang}, {and} \bibinfo{person}{Meng Wang}.} \bibinfo{year}{2020}\natexlab{}.
\newblock \showarticletitle{Lightgcn: Simplifying and powering graph convolution network for recommendation}. In \bibinfo{booktitle}{\emph{Proceedings of the 43rd International ACM SIGIR conference on research and development in Information Retrieval}}. \bibinfo{pages}{639--648}.
\newblock


\bibitem[Ji et~al\mbox{.}(2020)]%
        {ji2020dual}
\bibfield{author}{\bibinfo{person}{Shuyi Ji}, \bibinfo{person}{Yifan Feng}, \bibinfo{person}{Rongrong Ji}, \bibinfo{person}{Xibin Zhao}, \bibinfo{person}{Wanwan Tang}, {and} \bibinfo{person}{Yue Gao}.} \bibinfo{year}{2020}\natexlab{}.
\newblock \showarticletitle{Dual channel hypergraph collaborative filtering}. In \bibinfo{booktitle}{\emph{Proceedings of the 26th ACM SIGKDD international conference on knowledge discovery \& data mining}}. \bibinfo{pages}{2020--2029}.
\newblock


\bibitem[Jiang et~al\mbox{.}(2019)]%
        {jiang2019dynamic}
\bibfield{author}{\bibinfo{person}{Jianwen Jiang}, \bibinfo{person}{Yuxuan Wei}, \bibinfo{person}{Yifan Feng}, \bibinfo{person}{Jingxuan Cao}, {and} \bibinfo{person}{Yue Gao}.} \bibinfo{year}{2019}\natexlab{}.
\newblock \showarticletitle{Dynamic Hypergraph Neural Networks.}. In \bibinfo{booktitle}{\emph{IJCAI}}. \bibinfo{pages}{2635--2641}.
\newblock


\bibitem[Jin and Mont{\'u}far(2023)]%
        {jin2023implicit}
\bibfield{author}{\bibinfo{person}{Hui Jin} {and} \bibinfo{person}{Guido Mont{\'u}far}.} \bibinfo{year}{2023}\natexlab{}.
\newblock \showarticletitle{Implicit bias of gradient descent for mean squared error regression with two-layer wide neural networks}.
\newblock \bibinfo{journal}{\emph{Journal of Machine Learning Research}} \bibinfo{volume}{24}, \bibinfo{number}{137} (\bibinfo{year}{2023}), \bibinfo{pages}{1--97}.
\newblock


\bibitem[Kalofolias et~al\mbox{.}(2014)]%
        {kalofolias2014matrix}
\bibfield{author}{\bibinfo{person}{Vassilis Kalofolias}, \bibinfo{person}{Xavier Bresson}, \bibinfo{person}{Michael Bronstein}, {and} \bibinfo{person}{Pierre Vandergheynst}.} \bibinfo{year}{2014}\natexlab{}.
\newblock \showarticletitle{Matrix completion on graphs}.
\newblock \bibinfo{journal}{\emph{arXiv preprint arXiv:1408.1717}} (\bibinfo{year}{2014}).
\newblock


\bibitem[Kipf and Welling(2016)]%
        {kipf2016semi}
\bibfield{author}{\bibinfo{person}{Thomas~N Kipf} {and} \bibinfo{person}{Max Welling}.} \bibinfo{year}{2016}\natexlab{}.
\newblock \showarticletitle{Semi-supervised classification with graph convolutional networks}.
\newblock \bibinfo{journal}{\emph{arXiv preprint arXiv:1609.02907}} (\bibinfo{year}{2016}).
\newblock


\bibitem[Kumar et~al\mbox{.}(2020)]%
        {kumar2020link}
\bibfield{author}{\bibinfo{person}{Ajay Kumar}, \bibinfo{person}{Shashank~Sheshar Singh}, \bibinfo{person}{Kuldeep Singh}, {and} \bibinfo{person}{Bhaskar Biswas}.} \bibinfo{year}{2020}\natexlab{}.
\newblock \showarticletitle{Link prediction techniques, applications, and performance: A survey}.
\newblock \bibinfo{journal}{\emph{Physica A: Statistical Mechanics and its Applications}}  \bibinfo{volume}{553} (\bibinfo{year}{2020}), \bibinfo{pages}{124289}.
\newblock


\bibitem[Lan et~al\mbox{.}(2019)]%
        {lan2019albert}
\bibfield{author}{\bibinfo{person}{Zhenzhong Lan}, \bibinfo{person}{Mingda Chen}, \bibinfo{person}{Sebastian Goodman}, \bibinfo{person}{Kevin Gimpel}, \bibinfo{person}{Piyush Sharma}, {and} \bibinfo{person}{Radu Soricut}.} \bibinfo{year}{2019}\natexlab{}.
\newblock \showarticletitle{Albert: A lite bert for self-supervised learning of language representations}.
\newblock \bibinfo{journal}{\emph{arXiv preprint arXiv:1909.11942}} (\bibinfo{year}{2019}).
\newblock


\bibitem[Li et~al\mbox{.}(2023)]%
        {li2023sgccl}
\bibfield{author}{\bibinfo{person}{Boyu Li}, \bibinfo{person}{Ting Guo}, \bibinfo{person}{Xingquan Zhu}, \bibinfo{person}{Qian Li}, \bibinfo{person}{Yang Wang}, {and} \bibinfo{person}{Fang Chen}.} \bibinfo{year}{2023}\natexlab{}.
\newblock \showarticletitle{SGCCL: siamese graph contrastive consensus learning for personalized recommendation}. In \bibinfo{booktitle}{\emph{Proceedings of the Sixteenth ACM International Conference on Web Search and Data Mining}}. \bibinfo{pages}{589--597}.
\newblock


\bibitem[Li et~al\mbox{.}(2025)]%
        {li2025dual}
\bibfield{author}{\bibinfo{person}{Xiang Li}, \bibinfo{person}{Chaofan Fu}, \bibinfo{person}{Zhongying Zhao}, \bibinfo{person}{Guanjie Zheng}, \bibinfo{person}{Chao Huang}, \bibinfo{person}{Yanwei Yu}, {and} \bibinfo{person}{Junyu Dong}.} \bibinfo{year}{2025}\natexlab{}.
\newblock \showarticletitle{Dual-channel multiplex graph neural networks for recommendation}.
\newblock \bibinfo{journal}{\emph{IEEE Transactions on Knowledge and Data Engineering}} (\bibinfo{year}{2025}).
\newblock


\bibitem[Lin et~al\mbox{.}(2022)]%
        {lin2022improving}
\bibfield{author}{\bibinfo{person}{Zihan Lin}, \bibinfo{person}{Changxin Tian}, \bibinfo{person}{Yupeng Hou}, {and} \bibinfo{person}{Wayne~Xin Zhao}.} \bibinfo{year}{2022}\natexlab{}.
\newblock \showarticletitle{Improving graph collaborative filtering with neighborhood-enriched contrastive learning}. In \bibinfo{booktitle}{\emph{Proceedings of the ACM Web Conference 2022}}. \bibinfo{pages}{2320--2329}.
\newblock


\bibitem[Ma et~al\mbox{.}(2022)]%
        {ma2022crosscbr}
\bibfield{author}{\bibinfo{person}{Yunshan Ma}, \bibinfo{person}{Yingzhi He}, \bibinfo{person}{An Zhang}, \bibinfo{person}{Xiang Wang}, {and} \bibinfo{person}{Tat-Seng Chua}.} \bibinfo{year}{2022}\natexlab{}.
\newblock \showarticletitle{CrossCBR: cross-view contrastive learning for bundle recommendation}. In \bibinfo{booktitle}{\emph{Proceedings of the 28th ACM SIGKDD Conference on Knowledge Discovery and Data Mining}}. \bibinfo{pages}{1233--1241}.
\newblock


\bibitem[Mahadevan and Arock(2021)]%
        {mahadevan2021class}
\bibfield{author}{\bibinfo{person}{Anbazhagan Mahadevan} {and} \bibinfo{person}{Michael Arock}.} \bibinfo{year}{2021}\natexlab{}.
\newblock \showarticletitle{A class imbalance-aware review rating prediction using hybrid sampling and ensemble learning}.
\newblock \bibinfo{journal}{\emph{Multimedia Tools and Applications}}  \bibinfo{volume}{80} (\bibinfo{year}{2021}), \bibinfo{pages}{6911--6938}.
\newblock


\bibitem[Mart{\'\i}nez et~al\mbox{.}(2016)]%
        {martinez2016survey}
\bibfield{author}{\bibinfo{person}{V{\'\i}ctor Mart{\'\i}nez}, \bibinfo{person}{Fernando Berzal}, {and} \bibinfo{person}{Juan-Carlos Cubero}.} \bibinfo{year}{2016}\natexlab{}.
\newblock \showarticletitle{A survey of link prediction in complex networks}.
\newblock \bibinfo{journal}{\emph{ACM computing surveys (CSUR)}} \bibinfo{volume}{49}, \bibinfo{number}{4} (\bibinfo{year}{2016}), \bibinfo{pages}{1--33}.
\newblock


\bibitem[McNee et~al\mbox{.}(2006)]%
        {mcnee2006being}
\bibfield{author}{\bibinfo{person}{Sean~M McNee}, \bibinfo{person}{John Riedl}, {and} \bibinfo{person}{Joseph~A Konstan}.} \bibinfo{year}{2006}\natexlab{}.
\newblock \showarticletitle{Being accurate is not enough: how accuracy metrics have hurt recommender systems}. In \bibinfo{booktitle}{\emph{CHI'06 extended abstracts on Human factors in computing systems}}. \bibinfo{pages}{1097--1101}.
\newblock


\bibitem[Miller et~al\mbox{.}(2003)]%
        {miller2003movielens}
\bibfield{author}{\bibinfo{person}{Bradley~N Miller}, \bibinfo{person}{Istvan Albert}, \bibinfo{person}{Shyong~K Lam}, \bibinfo{person}{Joseph~A Konstan}, {and} \bibinfo{person}{John Riedl}.} \bibinfo{year}{2003}\natexlab{}.
\newblock \showarticletitle{Movielens unplugged: experiences with an occasionally connected recommender system}. In \bibinfo{booktitle}{\emph{Proceedings of the 8th international conference on Intelligent user interfaces}}. \bibinfo{pages}{263--266}.
\newblock


\bibitem[Min et~al\mbox{.}(2020)]%
        {min2020scattering}
\bibfield{author}{\bibinfo{person}{Yimeng Min}, \bibinfo{person}{Frederik Wenkel}, {and} \bibinfo{person}{Guy Wolf}.} \bibinfo{year}{2020}\natexlab{}.
\newblock \showarticletitle{Scattering gcn: Overcoming oversmoothness in graph convolutional networks}.
\newblock \bibinfo{journal}{\emph{Advances in neural information processing systems}}  \bibinfo{volume}{33} (\bibinfo{year}{2020}), \bibinfo{pages}{14498--14508}.
\newblock


\bibitem[Monti et~al\mbox{.}(2017)]%
        {monti2017geometric}
\bibfield{author}{\bibinfo{person}{Federico Monti}, \bibinfo{person}{Michael Bronstein}, {and} \bibinfo{person}{Xavier Bresson}.} \bibinfo{year}{2017}\natexlab{}.
\newblock \showarticletitle{Geometric matrix completion with recurrent multi-graph neural networks}.
\newblock \bibinfo{journal}{\emph{Advances in neural information processing systems}}  \bibinfo{volume}{30} (\bibinfo{year}{2017}).
\newblock


\bibitem[Ni et~al\mbox{.}(2019)]%
        {ni2019justifying}
\bibfield{author}{\bibinfo{person}{Jianmo Ni}, \bibinfo{person}{Jiacheng Li}, {and} \bibinfo{person}{Julian McAuley}.} \bibinfo{year}{2019}\natexlab{}.
\newblock \showarticletitle{Justifying recommendations using distantly-labeled reviews and fine-grained aspects}. In \bibinfo{booktitle}{\emph{Proceedings of the 2019 conference on empirical methods in natural language processing and the 9th international joint conference on natural language processing (EMNLP-IJCNLP)}}. \bibinfo{pages}{188--197}.
\newblock


\bibitem[Oord et~al\mbox{.}(2018)]%
        {oord2018representation}
\bibfield{author}{\bibinfo{person}{Aaron van~den Oord}, \bibinfo{person}{Yazhe Li}, {and} \bibinfo{person}{Oriol Vinyals}.} \bibinfo{year}{2018}\natexlab{}.
\newblock \showarticletitle{Representation learning with contrastive predictive coding}.
\newblock \bibinfo{journal}{\emph{arXiv preprint arXiv:1807.03748}} (\bibinfo{year}{2018}).
\newblock


\bibitem[Palomares et~al\mbox{.}(2018)]%
        {palomares2018multi}
\bibfield{author}{\bibinfo{person}{Ivan Palomares}, \bibinfo{person}{Fiona Browne}, {and} \bibinfo{person}{Peadar Davis}.} \bibinfo{year}{2018}\natexlab{}.
\newblock \showarticletitle{Multi-view fuzzy information fusion in collaborative filtering recommender systems: Application to the urban resilience domain}.
\newblock \bibinfo{journal}{\emph{Data \& Knowledge Engineering}}  \bibinfo{volume}{113} (\bibinfo{year}{2018}), \bibinfo{pages}{64--80}.
\newblock


\bibitem[Qi et~al\mbox{.}(2020)]%
        {qi2020mean}
\bibfield{author}{\bibinfo{person}{Jun Qi}, \bibinfo{person}{Jun Du}, \bibinfo{person}{Sabato~Marco Siniscalchi}, \bibinfo{person}{Xiaoli Ma}, {and} \bibinfo{person}{Chin-Hui Lee}.} \bibinfo{year}{2020}\natexlab{}.
\newblock \showarticletitle{On mean absolute error for deep neural network based vector-to-vector regression}.
\newblock \bibinfo{journal}{\emph{IEEE Signal Processing Letters}}  \bibinfo{volume}{27} (\bibinfo{year}{2020}), \bibinfo{pages}{1485--1489}.
\newblock


\bibitem[Rao et~al\mbox{.}(2015)]%
        {rao2015collaborative}
\bibfield{author}{\bibinfo{person}{Nikhil Rao}, \bibinfo{person}{Hsiang-Fu Yu}, \bibinfo{person}{Pradeep~K Ravikumar}, {and} \bibinfo{person}{Inderjit~S Dhillon}.} \bibinfo{year}{2015}\natexlab{}.
\newblock \showarticletitle{Collaborative filtering with graph information: Consistency and scalable methods}.
\newblock \bibinfo{journal}{\emph{Advances in neural information processing systems}}  \bibinfo{volume}{28} (\bibinfo{year}{2015}).
\newblock


\bibitem[Raza and Ding(2019)]%
        {raza2019progress}
\bibfield{author}{\bibinfo{person}{Shaina Raza} {and} \bibinfo{person}{Chen Ding}.} \bibinfo{year}{2019}\natexlab{}.
\newblock \showarticletitle{Progress in context-aware recommender systems—An overview}.
\newblock \bibinfo{journal}{\emph{Computer Science Review}}  \bibinfo{volume}{31} (\bibinfo{year}{2019}), \bibinfo{pages}{84--97}.
\newblock


\bibitem[Ren et~al\mbox{.}(2019)]%
        {ren2019heterogeneous}
\bibfield{author}{\bibinfo{person}{Yuxiang Ren}, \bibinfo{person}{Bo Liu}, \bibinfo{person}{Chao Huang}, \bibinfo{person}{Peng Dai}, \bibinfo{person}{Liefeng Bo}, {and} \bibinfo{person}{Jiawei Zhang}.} \bibinfo{year}{2019}\natexlab{}.
\newblock \showarticletitle{Heterogeneous deep graph infomax}.
\newblock \bibinfo{journal}{\emph{arXiv preprint arXiv:1911.08538}} (\bibinfo{year}{2019}).
\newblock


\bibitem[Ren et~al\mbox{.}(2022)]%
        {ren2022disentangled}
\bibfield{author}{\bibinfo{person}{Yuyang Ren}, \bibinfo{person}{Haonan Zhang}, \bibinfo{person}{Qi Li}, \bibinfo{person}{Luoyi Fu}, \bibinfo{person}{Jiaxin Ding}, \bibinfo{person}{Xinde Cao}, \bibinfo{person}{Xinbing Wang}, {and} \bibinfo{person}{Chenghu Zhou}.} \bibinfo{year}{2022}\natexlab{}.
\newblock \showarticletitle{Disentangled graph contrastive learning for review-based recommendation}.
\newblock \bibinfo{journal}{\emph{arXiv preprint arXiv:2209.01524}} (\bibinfo{year}{2022}).
\newblock


\bibitem[Ricci et~al\mbox{.}(2010)]%
        {ricci2010introduction}
\bibfield{author}{\bibinfo{person}{Francesco Ricci}, \bibinfo{person}{Lior Rokach}, {and} \bibinfo{person}{Bracha Shapira}.} \bibinfo{year}{2010}\natexlab{}.
\newblock \showarticletitle{Introduction to recommender systems handbook}.
\newblock In \bibinfo{booktitle}{\emph{Recommender systems handbook}}. \bibinfo{publisher}{Springer}, \bibinfo{pages}{1--35}.
\newblock


\bibitem[Seo et~al\mbox{.}(2017)]%
        {seo2017interpretable}
\bibfield{author}{\bibinfo{person}{Sungyong Seo}, \bibinfo{person}{Jing Huang}, \bibinfo{person}{Hao Yang}, {and} \bibinfo{person}{Yan Liu}.} \bibinfo{year}{2017}\natexlab{}.
\newblock \showarticletitle{Interpretable convolutional neural networks with dual local and global attention for review rating prediction}. In \bibinfo{booktitle}{\emph{Proceedings of the eleventh ACM conference on recommender systems}}. \bibinfo{pages}{297--305}.
\newblock


\bibitem[Shen et~al\mbox{.}(2021)]%
        {shen2021inductive}
\bibfield{author}{\bibinfo{person}{Wei Shen}, \bibinfo{person}{Chuheng Zhang}, \bibinfo{person}{Yun Tian}, \bibinfo{person}{Liang Zeng}, \bibinfo{person}{Xiaonan He}, \bibinfo{person}{Wanchun Dou}, {and} \bibinfo{person}{Xiaolong Xu}.} \bibinfo{year}{2021}\natexlab{}.
\newblock \showarticletitle{Inductive matrix completion using graph autoencoder}. In \bibinfo{booktitle}{\emph{Proceedings of the 30th ACM International Conference on Information \& Knowledge Management}}. \bibinfo{pages}{1609--1618}.
\newblock


\bibitem[Shi et~al\mbox{.}(2023)]%
        {shi2023deep}
\bibfield{author}{\bibinfo{person}{Xintong Shi}, \bibinfo{person}{Wenzhi Cao}, {and} \bibinfo{person}{Sebastian Raschka}.} \bibinfo{year}{2023}\natexlab{}.
\newblock \showarticletitle{Deep neural networks for rank-consistent ordinal regression based on conditional probabilities}.
\newblock \bibinfo{journal}{\emph{Pattern Analysis and Applications}} \bibinfo{volume}{26}, \bibinfo{number}{3} (\bibinfo{year}{2023}), \bibinfo{pages}{941--955}.
\newblock


\bibitem[Shuai et~al\mbox{.}(2023)]%
        {shuai2023topic}
\bibfield{author}{\bibinfo{person}{Jie Shuai}, \bibinfo{person}{Le Wu}, \bibinfo{person}{Kun Zhang}, \bibinfo{person}{Peijie Sun}, \bibinfo{person}{Richang Hong}, {and} \bibinfo{person}{Meng Wang}.} \bibinfo{year}{2023}\natexlab{}.
\newblock \showarticletitle{Topic-enhanced graph neural networks for extraction-based explainable recommendation}. In \bibinfo{booktitle}{\emph{Proceedings of the 46th International ACM SIGIR Conference on Research and Development in Information Retrieval}}. \bibinfo{pages}{1188--1197}.
\newblock


\bibitem[Shuai et~al\mbox{.}(2022)]%
        {shuai2022review}
\bibfield{author}{\bibinfo{person}{Jie Shuai}, \bibinfo{person}{Kun Zhang}, \bibinfo{person}{Le Wu}, \bibinfo{person}{Peijie Sun}, \bibinfo{person}{Richang Hong}, \bibinfo{person}{Meng Wang}, {and} \bibinfo{person}{Yong Li}.} \bibinfo{year}{2022}\natexlab{}.
\newblock \showarticletitle{A review-aware graph contrastive learning framework for recommendation}. In \bibinfo{booktitle}{\emph{Proceedings of the 45th International ACM SIGIR Conference on Research and Development in Information Retrieval}}. \bibinfo{pages}{1283--1293}.
\newblock


\bibitem[Su et~al\mbox{.}(2021)]%
        {su2021whitening}
\bibfield{author}{\bibinfo{person}{Jianlin Su}, \bibinfo{person}{Jiarun Cao}, \bibinfo{person}{Weijie Liu}, {and} \bibinfo{person}{Yangyiwen Ou}.} \bibinfo{year}{2021}\natexlab{}.
\newblock \showarticletitle{Whitening sentence representations for better semantics and faster retrieval}.
\newblock \bibinfo{journal}{\emph{arXiv preprint arXiv:2103.15316}} (\bibinfo{year}{2021}).
\newblock


\bibitem[Thakoor et~al\mbox{.}(2021)]%
        {thakoor2021bootstrapped}
\bibfield{author}{\bibinfo{person}{Shantanu Thakoor}, \bibinfo{person}{Corentin Tallec}, \bibinfo{person}{Mohammad~Gheshlaghi Azar}, \bibinfo{person}{R{\'e}mi Munos}, \bibinfo{person}{Petar Veli{\v{c}}kovi{\'c}}, {and} \bibinfo{person}{Michal Valko}.} \bibinfo{year}{2021}\natexlab{}.
\newblock \showarticletitle{Bootstrapped representation learning on graphs}. In \bibinfo{booktitle}{\emph{ICLR 2021 Workshop on Geometrical and Topological Representation Learning}}.
\newblock


\bibitem[Veli{\v{c}}kovi{\'c} et~al\mbox{.}(2017)]%
        {velivckovic2017graph}
\bibfield{author}{\bibinfo{person}{Petar Veli{\v{c}}kovi{\'c}}, \bibinfo{person}{Guillem Cucurull}, \bibinfo{person}{Arantxa Casanova}, \bibinfo{person}{Adriana Romero}, \bibinfo{person}{Pietro Lio}, {and} \bibinfo{person}{Yoshua Bengio}.} \bibinfo{year}{2017}\natexlab{}.
\newblock \showarticletitle{Graph attention networks}.
\newblock \bibinfo{journal}{\emph{arXiv preprint arXiv:1710.10903}} (\bibinfo{year}{2017}).
\newblock


\bibitem[Veli{\v{c}}kovi{\'c} et~al\mbox{.}(2018)]%
        {velivckovic2018deep}
\bibfield{author}{\bibinfo{person}{Petar Veli{\v{c}}kovi{\'c}}, \bibinfo{person}{William Fedus}, \bibinfo{person}{William~L Hamilton}, \bibinfo{person}{Pietro Li{\`o}}, \bibinfo{person}{Yoshua Bengio}, {and} \bibinfo{person}{R~Devon Hjelm}.} \bibinfo{year}{2018}\natexlab{}.
\newblock \showarticletitle{Deep graph infomax}.
\newblock \bibinfo{journal}{\emph{arXiv preprint arXiv:1809.10341}} (\bibinfo{year}{2018}).
\newblock


\bibitem[Wang et~al\mbox{.}(2024)]%
        {wang2024enhanced}
\bibfield{author}{\bibinfo{person}{Ke Wang}, \bibinfo{person}{Yanmin Zhu}, \bibinfo{person}{Tianzi Zang}, \bibinfo{person}{Chunyang Wang}, {and} \bibinfo{person}{Mengyuan Jing}.} \bibinfo{year}{2024}\natexlab{}.
\newblock \showarticletitle{Review-enhanced hierarchical contrastive learning for recommendation}. In \bibinfo{booktitle}{\emph{Proceedings of the AAAI Conference on Artificial Intelligence}}, Vol.~\bibinfo{volume}{38}. \bibinfo{pages}{9107--9115}.
\newblock


\bibitem[Wang et~al\mbox{.}({[n.\,d.]})]%
        {wangmulti}
\bibfield{author}{\bibinfo{person}{Ke Wang}, \bibinfo{person}{Yanmin Zhu}, \bibinfo{person}{Tianzi Zang}, \bibinfo{person}{Chunyang Wang}, \bibinfo{person}{Kuan Liu}, {and} \bibinfo{person}{Peibo Ma}.} \bibinfo{year}{[n.\,d.]}\natexlab{}.
\newblock \showarticletitle{Multi-aspect Graph Contrastive Learning for Review-enhanced Recommendation}.
\newblock \bibinfo{journal}{\emph{ACM Transactions on Information Systems}} (\bibinfo{year}{[n.\,d.]}).
\newblock


\bibitem[Wang et~al\mbox{.}(2023)]%
        {wang2023multi}
\bibfield{author}{\bibinfo{person}{Ke Wang}, \bibinfo{person}{Yanmin Zhu}, \bibinfo{person}{Tianzi Zang}, \bibinfo{person}{Chunyang Wang}, \bibinfo{person}{Kuan Liu}, {and} \bibinfo{person}{Peibo Ma}.} \bibinfo{year}{2023}\natexlab{}.
\newblock \showarticletitle{Multi-aspect Graph Contrastive Learning for Review-enhanced Recommendation}.
\newblock \bibinfo{journal}{\emph{ACM Transactions on Information Systems}} \bibinfo{volume}{42}, \bibinfo{number}{2} (\bibinfo{year}{2023}), \bibinfo{pages}{1--29}.
\newblock


\bibitem[Wang(2019)]%
        {wang2019deep}
\bibfield{author}{\bibinfo{person}{Minjie~Yu Wang}.} \bibinfo{year}{2019}\natexlab{}.
\newblock \showarticletitle{Deep graph library: Towards efficient and scalable deep learning on graphs}. In \bibinfo{booktitle}{\emph{ICLR workshop on representation learning on graphs and manifolds}}.
\newblock


\bibitem[Wang et~al\mbox{.}(2019)]%
        {wang2019neural}
\bibfield{author}{\bibinfo{person}{Xiang Wang}, \bibinfo{person}{Xiangnan He}, \bibinfo{person}{Meng Wang}, \bibinfo{person}{Fuli Feng}, {and} \bibinfo{person}{Tat-Seng Chua}.} \bibinfo{year}{2019}\natexlab{}.
\newblock \showarticletitle{Neural graph collaborative filtering}. In \bibinfo{booktitle}{\emph{Proceedings of the 42nd international ACM SIGIR conference on Research and development in Information Retrieval}}. \bibinfo{pages}{165--174}.
\newblock


\bibitem[Wei et~al\mbox{.}(2022)]%
        {wei2022dynamic}
\bibfield{author}{\bibinfo{person}{Chunyu Wei}, \bibinfo{person}{Jian Liang}, \bibinfo{person}{Bing Bai}, {and} \bibinfo{person}{Di Liu}.} \bibinfo{year}{2022}\natexlab{}.
\newblock \showarticletitle{Dynamic Hypergraph Learning for Collaborative Filtering}. In \bibinfo{booktitle}{\emph{Proceedings of the 31st ACM International Conference on Information \& Knowledge Management}}. \bibinfo{pages}{2108--2117}.
\newblock


\bibitem[Wei et~al\mbox{.}(2021)]%
        {wei2021towards}
\bibfield{author}{\bibinfo{person}{Tong Wei}, \bibinfo{person}{Wei-Wei Tu}, \bibinfo{person}{Yu-Feng Li}, {and} \bibinfo{person}{Guo-Ping Yang}.} \bibinfo{year}{2021}\natexlab{}.
\newblock \showarticletitle{Towards robust prediction on tail labels}. In \bibinfo{booktitle}{\emph{Proceedings of the 27th ACM SIGKDD Conference on Knowledge Discovery \& Data Mining}}. \bibinfo{pages}{1812--1820}.
\newblock


\bibitem[Wu et~al\mbox{.}(2019)]%
        {wu2019reviews}
\bibfield{author}{\bibinfo{person}{Chuhan Wu}, \bibinfo{person}{Fangzhao Wu}, \bibinfo{person}{Tao Qi}, \bibinfo{person}{Suyu Ge}, \bibinfo{person}{Yongfeng Huang}, {and} \bibinfo{person}{Xing Xie}.} \bibinfo{year}{2019}\natexlab{}.
\newblock \showarticletitle{Reviews meet graphs: enhancing user and item representations for recommendation with hierarchical attentive graph neural network}. In \bibinfo{booktitle}{\emph{Proceedings of the 2019 Conference on Empirical Methods in Natural Language Processing and the 9th International Joint Conference on Natural Language Processing (EMNLP-IJCNLP)}}. \bibinfo{pages}{4884--4893}.
\newblock


\bibitem[Wu et~al\mbox{.}(2021a)]%
        {wu2021self}
\bibfield{author}{\bibinfo{person}{Jiancan Wu}, \bibinfo{person}{Xiang Wang}, \bibinfo{person}{Fuli Feng}, \bibinfo{person}{Xiangnan He}, \bibinfo{person}{Liang Chen}, \bibinfo{person}{Jianxun Lian}, {and} \bibinfo{person}{Xing Xie}.} \bibinfo{year}{2021}\natexlab{a}.
\newblock \showarticletitle{Self-supervised graph learning for recommendation}. In \bibinfo{booktitle}{\emph{Proceedings of the 44th international ACM SIGIR conference on research and development in information retrieval}}. \bibinfo{pages}{726--735}.
\newblock


\bibitem[Wu et~al\mbox{.}(2021b)]%
        {wu2021towards}
\bibfield{author}{\bibinfo{person}{Qitian Wu}, \bibinfo{person}{Hengrui Zhang}, \bibinfo{person}{Xiaofeng Gao}, \bibinfo{person}{Junchi Yan}, {and} \bibinfo{person}{Hongyuan Zha}.} \bibinfo{year}{2021}\natexlab{b}.
\newblock \showarticletitle{Towards open-world recommendation: An inductive model-based collaborative filtering approach}. In \bibinfo{booktitle}{\emph{International Conference on Machine Learning}}. PMLR, \bibinfo{pages}{11329--11339}.
\newblock


\bibitem[Wu and Huang(2022)]%
        {wu2022gumbel}
\bibfield{author}{\bibinfo{person}{Yuexin Wu} {and} \bibinfo{person}{Xiaolei Huang}.} \bibinfo{year}{2022}\natexlab{}.
\newblock \showarticletitle{A Gumbel-based Rating Prediction Framework for Imbalanced Recommendation}. In \bibinfo{booktitle}{\emph{Proceedings of the 31st ACM International Conference on Information \& Knowledge Management}}. \bibinfo{pages}{2199--2209}.
\newblock


\bibitem[Xia et~al\mbox{.}(2020)]%
        {xia2020multiplex}
\bibfield{author}{\bibinfo{person}{Lianghao Xia}, \bibinfo{person}{Chao Huang}, \bibinfo{person}{Yong Xu}, \bibinfo{person}{Peng Dai}, \bibinfo{person}{Bo Zhang}, {and} \bibinfo{person}{Liefeng Bo}.} \bibinfo{year}{2020}\natexlab{}.
\newblock \showarticletitle{Multiplex behavioral relation learning for recommendation via memory augmented transformer network}. In \bibinfo{booktitle}{\emph{Proceedings of the 43rd international ACM SIGIR conference on research and development in information retrieval}}. \bibinfo{pages}{2397--2406}.
\newblock


\bibitem[Xia et~al\mbox{.}(2022)]%
        {xia2022hypergraph}
\bibfield{author}{\bibinfo{person}{Lianghao Xia}, \bibinfo{person}{Chao Huang}, \bibinfo{person}{Yong Xu}, \bibinfo{person}{Jiashu Zhao}, \bibinfo{person}{Dawei Yin}, {and} \bibinfo{person}{Jimmy Huang}.} \bibinfo{year}{2022}\natexlab{}.
\newblock \showarticletitle{Hypergraph contrastive collaborative filtering}. In \bibinfo{booktitle}{\emph{Proceedings of the 45th International ACM SIGIR conference on research and development in information retrieval}}. \bibinfo{pages}{70--79}.
\newblock


\bibitem[Xu et~al\mbox{.}(2013)]%
        {xu2013speedup}
\bibfield{author}{\bibinfo{person}{Miao Xu}, \bibinfo{person}{Rong Jin}, {and} \bibinfo{person}{Zhi-Hua Zhou}.} \bibinfo{year}{2013}\natexlab{}.
\newblock \showarticletitle{Speedup matrix completion with side information: Application to multi-label learning}.
\newblock \bibinfo{journal}{\emph{Advances in neural information processing systems}}  \bibinfo{volume}{26} (\bibinfo{year}{2013}).
\newblock


\bibitem[Xue et~al\mbox{.}(2021)]%
        {xue2021multiplex}
\bibfield{author}{\bibinfo{person}{Hansheng Xue}, \bibinfo{person}{Luwei Yang}, \bibinfo{person}{Vaibhav Rajan}, \bibinfo{person}{Wen Jiang}, \bibinfo{person}{Yi Wei}, {and} \bibinfo{person}{Yu Lin}.} \bibinfo{year}{2021}\natexlab{}.
\newblock \showarticletitle{Multiplex bipartite network embedding using dual hypergraph convolutional networks}. In \bibinfo{booktitle}{\emph{Proceedings of the Web Conference 2021}}. \bibinfo{pages}{1649--1660}.
\newblock


\bibitem[Yang et~al\mbox{.}(2019)]%
        {yang2019revisiting}
\bibfield{author}{\bibinfo{person}{Dingqi Yang}, \bibinfo{person}{Bingqing Qu}, \bibinfo{person}{Jie Yang}, {and} \bibinfo{person}{Philippe Cudre-Mauroux}.} \bibinfo{year}{2019}\natexlab{}.
\newblock \showarticletitle{Revisiting user mobility and social relationships in lbsns: a hypergraph embedding approach}. In \bibinfo{booktitle}{\emph{The world wide web conference}}. \bibinfo{pages}{2147--2157}.
\newblock


\bibitem[Yang et~al\mbox{.}(2023)]%
        {yang2023generative}
\bibfield{author}{\bibinfo{person}{Yonghui Yang}, \bibinfo{person}{Zhengwei Wu}, \bibinfo{person}{Le Wu}, \bibinfo{person}{Kun Zhang}, \bibinfo{person}{Richang Hong}, \bibinfo{person}{Zhiqiang Zhang}, \bibinfo{person}{Jun Zhou}, {and} \bibinfo{person}{Meng Wang}.} \bibinfo{year}{2023}\natexlab{}.
\newblock \showarticletitle{Generative-contrastive graph learning for recommendation}. In \bibinfo{booktitle}{\emph{Proceedings of the 46th international ACM SIGIR Conference on Research and Development in Information Retrieval}}. \bibinfo{pages}{1117--1126}.
\newblock


\bibitem[Ye et~al\mbox{.}(2023)]%
        {ye2023towards}
\bibfield{author}{\bibinfo{person}{Haibo Ye}, \bibinfo{person}{Xinjie Li}, \bibinfo{person}{Yuan Yao}, {and} \bibinfo{person}{Hanghang Tong}.} \bibinfo{year}{2023}\natexlab{}.
\newblock \showarticletitle{Towards robust neural graph collaborative filtering via structure denoising and embedding perturbation}.
\newblock \bibinfo{journal}{\emph{ACM Transactions on Information Systems}} \bibinfo{volume}{41}, \bibinfo{number}{3} (\bibinfo{year}{2023}), \bibinfo{pages}{1--28}.
\newblock


\bibitem[Ying et~al\mbox{.}(2018)]%
        {ying2018graph}
\bibfield{author}{\bibinfo{person}{Rex Ying}, \bibinfo{person}{Ruining He}, \bibinfo{person}{Kaifeng Chen}, \bibinfo{person}{Pong Eksombatchai}, \bibinfo{person}{William~L Hamilton}, {and} \bibinfo{person}{Jure Leskovec}.} \bibinfo{year}{2018}\natexlab{}.
\newblock \showarticletitle{Graph convolutional neural networks for web-scale recommender systems}. In \bibinfo{booktitle}{\emph{Proceedings of the 24th ACM SIGKDD international conference on knowledge discovery \& data mining}}. \bibinfo{pages}{974--983}.
\newblock


\bibitem[You et~al\mbox{.}(2020)]%
        {you2020graph}
\bibfield{author}{\bibinfo{person}{Yuning You}, \bibinfo{person}{Tianlong Chen}, \bibinfo{person}{Yongduo Sui}, \bibinfo{person}{Ting Chen}, \bibinfo{person}{Zhangyang Wang}, {and} \bibinfo{person}{Yang Shen}.} \bibinfo{year}{2020}\natexlab{}.
\newblock \showarticletitle{Graph contrastive learning with augmentations}.
\newblock \bibinfo{journal}{\emph{Advances in neural information processing systems}}  \bibinfo{volume}{33} (\bibinfo{year}{2020}), \bibinfo{pages}{5812--5823}.
\newblock


\bibitem[Yu et~al\mbox{.}(2023)]%
        {yu2023xsimgcl}
\bibfield{author}{\bibinfo{person}{Junliang Yu}, \bibinfo{person}{Xin Xia}, \bibinfo{person}{Tong Chen}, \bibinfo{person}{Lizhen Cui}, \bibinfo{person}{Nguyen Quoc~Viet Hung}, {and} \bibinfo{person}{Hongzhi Yin}.} \bibinfo{year}{2023}\natexlab{}.
\newblock \showarticletitle{XSimGCL: Towards extremely simple graph contrastive learning for recommendation}.
\newblock \bibinfo{journal}{\emph{IEEE Transactions on Knowledge and Data Engineering}} (\bibinfo{year}{2023}).
\newblock


\bibitem[Yu et~al\mbox{.}(2021)]%
        {yu2021self}
\bibfield{author}{\bibinfo{person}{Junliang Yu}, \bibinfo{person}{Hongzhi Yin}, \bibinfo{person}{Jundong Li}, \bibinfo{person}{Qinyong Wang}, \bibinfo{person}{Nguyen Quoc~Viet Hung}, {and} \bibinfo{person}{Xiangliang Zhang}.} \bibinfo{year}{2021}\natexlab{}.
\newblock \showarticletitle{Self-supervised multi-channel hypergraph convolutional network for social recommendation}. In \bibinfo{booktitle}{\emph{Proceedings of the web conference 2021}}. \bibinfo{pages}{413--424}.
\newblock


\bibitem[Yu et~al\mbox{.}(2022)]%
        {yu2022graph}
\bibfield{author}{\bibinfo{person}{Junliang Yu}, \bibinfo{person}{Hongzhi Yin}, \bibinfo{person}{Xin Xia}, \bibinfo{person}{Tong Chen}, \bibinfo{person}{Lizhen Cui}, {and} \bibinfo{person}{Quoc Viet~Hung Nguyen}.} \bibinfo{year}{2022}\natexlab{}.
\newblock \showarticletitle{Are graph augmentations necessary? simple graph contrastive learning for recommendation}. In \bibinfo{booktitle}{\emph{Proceedings of the 45th international ACM SIGIR conference on research and development in information retrieval}}. \bibinfo{pages}{1294--1303}.
\newblock


\bibitem[Yuan et~al\mbox{.}(2021)]%
        {yuan2021multimodal}
\bibfield{author}{\bibinfo{person}{Xin Yuan}, \bibinfo{person}{Zhe Lin}, \bibinfo{person}{Jason Kuen}, \bibinfo{person}{Jianming Zhang}, \bibinfo{person}{Yilin Wang}, \bibinfo{person}{Michael Maire}, \bibinfo{person}{Ajinkya Kale}, {and} \bibinfo{person}{Baldo Faieta}.} \bibinfo{year}{2021}\natexlab{}.
\newblock \showarticletitle{Multimodal contrastive training for visual representation learning}. In \bibinfo{booktitle}{\emph{Proceedings of the IEEE/CVF Conference on Computer Vision and Pattern Recognition}}. \bibinfo{pages}{6995--7004}.
\newblock


\bibitem[Zhang et~al\mbox{.}(2022)]%
        {zhang2022geometric}
\bibfield{author}{\bibinfo{person}{Chengkun Zhang}, \bibinfo{person}{Hongxu Chen}, \bibinfo{person}{Sixiao Zhang}, \bibinfo{person}{Guandong Xu}, {and} \bibinfo{person}{Junbin Gao}.} \bibinfo{year}{2022}\natexlab{}.
\newblock \showarticletitle{Geometric Inductive Matrix Completion: A Hyperbolic Approach with Unified Message Passing}. In \bibinfo{booktitle}{\emph{Proceedings of the Fifteenth ACM International Conference on Web Search and Data Mining}}. \bibinfo{pages}{1337--1346}.
\newblock


\bibitem[Zhang et~al\mbox{.}(2019)]%
        {zhang2019star}
\bibfield{author}{\bibinfo{person}{Jiani Zhang}, \bibinfo{person}{Xingjian Shi}, \bibinfo{person}{Shenglin Zhao}, {and} \bibinfo{person}{Irwin King}.} \bibinfo{year}{2019}\natexlab{}.
\newblock \showarticletitle{Star-gcn: Stacked and reconstructed graph convolutional networks for recommender systems}.
\newblock \bibinfo{journal}{\emph{arXiv preprint arXiv:1905.13129}} (\bibinfo{year}{2019}).
\newblock


\bibitem[Zhang and Chen(2019)]%
        {zhang2019inductive}
\bibfield{author}{\bibinfo{person}{Muhan Zhang} {and} \bibinfo{person}{Yixin Chen}.} \bibinfo{year}{2019}\natexlab{}.
\newblock \showarticletitle{Inductive matrix completion based on graph neural networks}.
\newblock \bibinfo{journal}{\emph{arXiv preprint arXiv:1904.12058}} (\bibinfo{year}{2019}).
\newblock


\bibitem[Zhang et~al\mbox{.}(2018)]%
        {zhang2018dynamic}
\bibfield{author}{\bibinfo{person}{Zizhao Zhang}, \bibinfo{person}{Haojie Lin}, \bibinfo{person}{Yue Gao}, {and} \bibinfo{person}{KLISS BNRist}.} \bibinfo{year}{2018}\natexlab{}.
\newblock \showarticletitle{Dynamic hypergraph structure learning.}. In \bibinfo{booktitle}{\emph{IJCAI}}. \bibinfo{pages}{3162--3169}.
\newblock


\bibitem[Zhao et~al\mbox{.}(2020)]%
        {zhao2020improving}
\bibfield{author}{\bibinfo{person}{Xing Zhao}, \bibinfo{person}{Ziwei Zhu}, \bibinfo{person}{Yin Zhang}, {and} \bibinfo{person}{James Caverlee}.} \bibinfo{year}{2020}\natexlab{}.
\newblock \showarticletitle{Improving the estimation of tail ratings in recommender system with multi-latent representations}. In \bibinfo{booktitle}{\emph{Proceedings of the 13th International Conference on Web Search and Data Mining}}. \bibinfo{pages}{762--770}.
\newblock


\bibitem[Zhou et~al\mbox{.}(2006)]%
        {zhou2006learning}
\bibfield{author}{\bibinfo{person}{Dengyong Zhou}, \bibinfo{person}{Jiayuan Huang}, {and} \bibinfo{person}{Bernhard Sch{\"o}lkopf}.} \bibinfo{year}{2006}\natexlab{}.
\newblock \showarticletitle{Learning with hypergraphs: Clustering, classification, and embedding}.
\newblock \bibinfo{journal}{\emph{Advances in neural information processing systems}}  \bibinfo{volume}{19} (\bibinfo{year}{2006}).
\newblock


\bibitem[Zhu et~al\mbox{.}(2020)]%
        {zhu2020deep}
\bibfield{author}{\bibinfo{person}{Yanqiao Zhu}, \bibinfo{person}{Yichen Xu}, \bibinfo{person}{Feng Yu}, \bibinfo{person}{Qiang Liu}, \bibinfo{person}{Shu Wu}, {and} \bibinfo{person}{Liang Wang}.} \bibinfo{year}{2020}\natexlab{}.
\newblock \showarticletitle{Deep graph contrastive representation learning}.
\newblock \bibinfo{journal}{\emph{arXiv preprint arXiv:2006.04131}} (\bibinfo{year}{2020}).
\newblock


\end{thebibliography}

\end{document}